\documentclass[journal]{IEEEtran}

\usepackage{cite}
\usepackage{times}
\usepackage{epsfig}
\usepackage{graphicx}
\usepackage{amsmath}
\usepackage{amssymb}
\usepackage{booktabs}
\usepackage{multirow}
\usepackage{verbatim}
\usepackage{float}
\usepackage{amsmath}
\begin{document}
%
\title{A New Dataset, Poisson GAN and AquaNet for Underwater Object Grabbing}
\author{Chongwei Liu, Zhihui Wang, Shijie Wang, Tao Tang, Yulong Tao, Caifei Yang, Haojie Li, Xing Liu, and Xin Fan
\thanks{Copyright $\copyright$ 2021 IEEE. Personal use of this material is permitted. However, permission to use this material for any other purposes must be obtained from the IEEE by sending a request to pubs-permissions@ieee.org.}
\thanks{C. Liu, Z. Wang, S. Wang, T. Tang, C. Yang, H. Li, X. Liu, and X. Fan are with DUT-RU International School of Information Science \& Engineering, Dalian University of Technology, Liaoning. }

\thanks{Y. Tao is with the Department of LIESMARS, The Wuhan University, Wuhan. (Corresponding author: Z. Wang. E-mail: zhwang@dlut.edu.cn)}

}

\maketitle

\begin{abstract}
To boost the object grabbing capability of underwater robots for open-sea farming, we propose a new dataset (UDD) consisting of three categories (seacucumber, seaurchin, and scallop) with 2,227 images. To the best of our knowledge, it is the first 4K HD dataset collected in a real open-sea farm. We also propose a novel Poisson-blending Generative Adversarial Network (Poisson GAN) and an efficient object detection network (AquaNet) to address two common issues within related datasets: the class-imbalance problem and the problem of mass small object, respectively. 
Specifically, Poisson GAN combines Poisson blending into its generator and employs a new loss called Dual Restriction loss (DR loss), which supervises both implicit space features and image-level features during training to generate more realistic images.
 By utilizing Poisson GAN, objects of minority class like seacucumber or scallop could be added into an image naturally and annotated automatically, which could increase the loss of minority classes during training detectors to eliminate the class-imbalance problem;
AquaNet is a high-efficiency detector to address the problem of detecting mass small objects from cloudy underwater pictures. Within it, we design two efficient components: a depth-wise-convolution-based Multi-scale Contextual Features Fusion (MFF) block and a Multi-scale Blursampling (MBP) module to reduce the parameters of the network to 1.3 million. Both two components could provide multi-scale features of small objects under a short backbone configuration without any loss of accuracy. In addition, we construct a large-scale augmented dataset (AUDD) and a pre-training dataset via Poisson GAN from UDD. Extensive experiments show the effectiveness of the proposed Poisson GAN, AquaNet, UDD, AUDD, and pre-training dataset.
\end{abstract}


\IEEEpeerreviewmaketitle

\section{Introduction}
Underwater robot picking means using robots to grab mariculture organisms like seacucumber, seaurchin, or scallop in an open-sea farm automatically. By the end of 2019, the total output of planted seacucumber was about 170,000 tons with an output value of \$ 4.9 billion in China. However, divers are still the main force in harvesting sea cucumbers or other creatures. Developing underwater robot picking could improve the efficiency and protect divers from health problems caused by the long time diving. Underwater object detection is a key step in underwater robot automatic picking task. In recent years, due to the superior feature representation ability of deep CNNs \cite{8489974,Krizhevsky2012ImageNet} and the availability of large datasets (\emph{e.g.}, MS COCO \cite{Belongie2014}), general object detection achieved remarkable success. However, less progress has been made in underwater object detection because there are 
three major tough challenges to overcome:
\begin{figure}[!t]
\begin{center}
\includegraphics[width=1\linewidth]{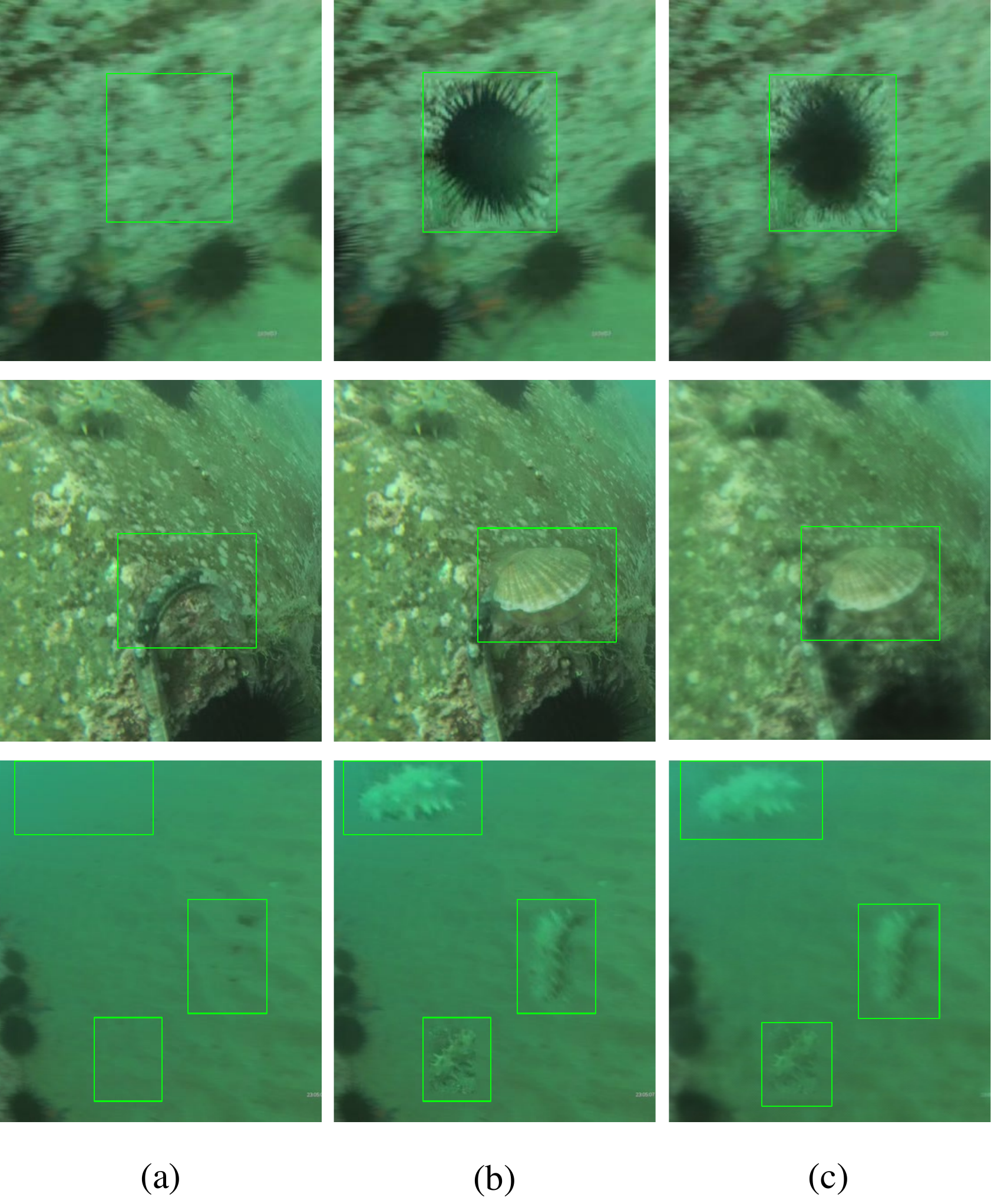}
\end{center}
   \caption{Images generated by Poisson GAN: (a) original images, (b) images produced by Poisson blending phase and (c) images produced by learning phase. The details are high-lighted in green boxes (added for visualization).}
\label{comparison3}
\end{figure}
\begin{figure*}[!t]
\begin{center}
\includegraphics[width=1\linewidth]{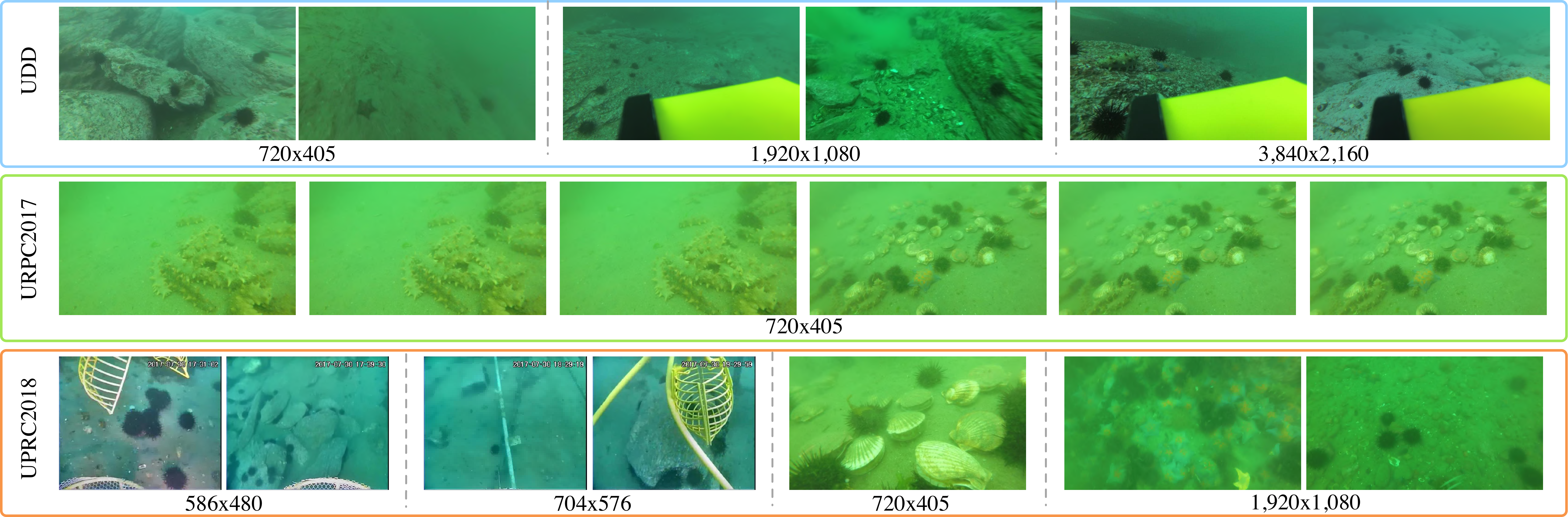}
\end{center}
   \caption{Images in UDD, URPC2017, and URPC2018. Our UDD covers a variety of sea farm scenes and has native high-resolution images (up to 4K). However, images in URPC2017 are highly similar and images in URPC2018 are low quality (zoom in to get a better view).}
\label{samples}
\end{figure*}
\begin{itemize}
	\item {\bf Lack of a benchmark dataset for research.}
Because the cost of underwater image collection is much more expensive than ordinary  cases, only a few datasets (\emph{e.g.}, URPC2017 and URPC2018) have been released by the Underwater Robot Picking Contest (URPC)$\protect\footnote{Underwater Robot Picking Contest: http://en.cnurpc.org. A video that includes the robot picking scene is also provided in https://haokan.baidu.com/v?pd=wisenatural\&vid=8062577669004183587 for audience to understand this field better.}$. 
However, the image quality and the acquisition environment are not sufficient for the later research, and the testing sets or comparable results of URPC datasets are not available.
   \item {\bf Class-imbalance and mass small objects.}
Due to the great differences in economic values among sea foods, the breed quantity of different aquatic products is different. Hence, the class-imbalance problem$\protect\footnote{The imbalance of different class instance numbers within one dataset, so the trained detectors usually perform badly in little number classes.}$ is prominent in related datasets. Meanwhile, 
most objects in the underwater image tend to be small. Both of them aggravate the difficulty of underwater objects detection.
    \item {\bf Efficiency requirements.}
For underwater robot grabbing, the image-based detector is a key part of a grabbing system. 
The computing resources and power of the on-board grasping system are limited, and other operations such as ranging and control also occupy the computing resources. Hence, detectors have to balance the accuracy and speed.

\end{itemize}

Considering the above challenges, we make efforts along three directions respectively:

First, we propose a 4K HD dataset and corresponding comparable results for further research. The underwater open-sea farm object detection dataset (UDD) contains 3 categories (seacucumber, seaurchin, and scallop) with 2,227 images. To simulate the real underwater grabbing environment, underwater robots and divers with a 4K HD camera were employed to capture images in a real open-sea farm. Then Poisson GAN expands UDD into 2 datasets: AUDD including about 18,000 images and pre-training dataset including 590,000 images. 
AUDD is an extension of UDD and pre-training dataset could make a detector better adapted to the underwater grabbing environment. When training detectors, we train them on AUDD with parameters pre-trained on the pre-training dataset. This strategy improves detectors significantly over  ImageNet pre-trained models or randomly initialized models on UDD. 

Second, to address the class-imbalance problem, we propose Poisson GAN. Compared with existing GAN-based data augmentation methods \cite{8962219,Deng2017Image,Wei2018Person}, we embed Poisson blending \cite{Rez2003Poisson} into the generator as the Poisson blending phase to change the number, position, even size of objects in an image and employ a network as the learning phase to generate realistic images. Fig. \ref{comparison3} shows images processed by each phase. 
Furthermore, we design a Dual Restriction loss (DR loss) which could supervise the implicit space features and the image-level features at the same time to generate more realistic images. Our proposed Poisson GAN can solve the class-imbalance problem in underwater object detection more effectively while others can only change image styles. 

Third, we design AquaNet to meet the requirement of accurately and efficiently detecting mass small objects$\protect\footnote{The phrase "mass small objects" refers a large number of small objects.}$ from cloudy underwater images. Specifically, two lightweight components are proposed: a pooling layer called Multi-scale Blursampling (MBP) module and a basic block called Multi-scale Contextual Features Fusion (MFF) block based on the depth-separable inverted bottleneck block \cite{Sandler2018MobileNetV2}. Both of them can extract and fuse multi-scale features at the same time, thus making AquaNet achieve high detection efficiency and promising accuracy. Code and datasets will be released on https://github.com/chongweiliu soon.

In summary, the contributions of this paper can be listed as follows.
\begin{itemize}
	\item A 4K HD benchmark (UDD) and corresponding comparable results are proposed for further research. Via Poisson GAN, a large-scale augmented underwater sea farm object detection dataset (AUDD) and a large-scale pre-training dataset are also proposed for underwater robot grabbing.

	\item By combining conventional and deep learning methods, we propose Poisson GAN to change the number, position, even size of objects in an image for effective underwater image augmentation, while existing GANs can barely do.
   
    \item A lightweight AquaNet with two efficient modules are designed to meet the efficiency requirements of the underwater robot grabbing without any loss on accuracy.
\end{itemize}

\section{Related Work}
This section briefly reviews the relevant fields, including the introduction of background and related datasets, data augmentation with GAN, as well as underwater object detection.
\subsection{Background and Related Datasets}
Underwater object detection for robot picking is first proposed in the target recognition track of Underwater Robot Picking Contest 2017 (URPC2017) which aims to promote the development of theory, technology, and industry of the underwater agile robot and fill the blank of the grabbing task of the  underwater agile robot. The competition sets up a target recognition track, a fixed-point grasping track, and an autonomous grasping track. The target recognition track concentrates on the algorithm and technology of underwater target recognition and detection for seacucumber, seaurchin, and scallop. 

So far, there are 2 available datasets:
\subsubsection{URPC2017}It contains 17,655 images for training and 985 images for testing and the resolution of all the images is 720$\times$405. However, all images are taken from 6 videos at an interval of 10 frames. All the videos were filmed in an artificial simulated environment and pictures from the same video look almost identical as shown in Fig. \ref {samples}. 
\begin{figure}[!t]
\begin{center}
\includegraphics[width=1\linewidth]{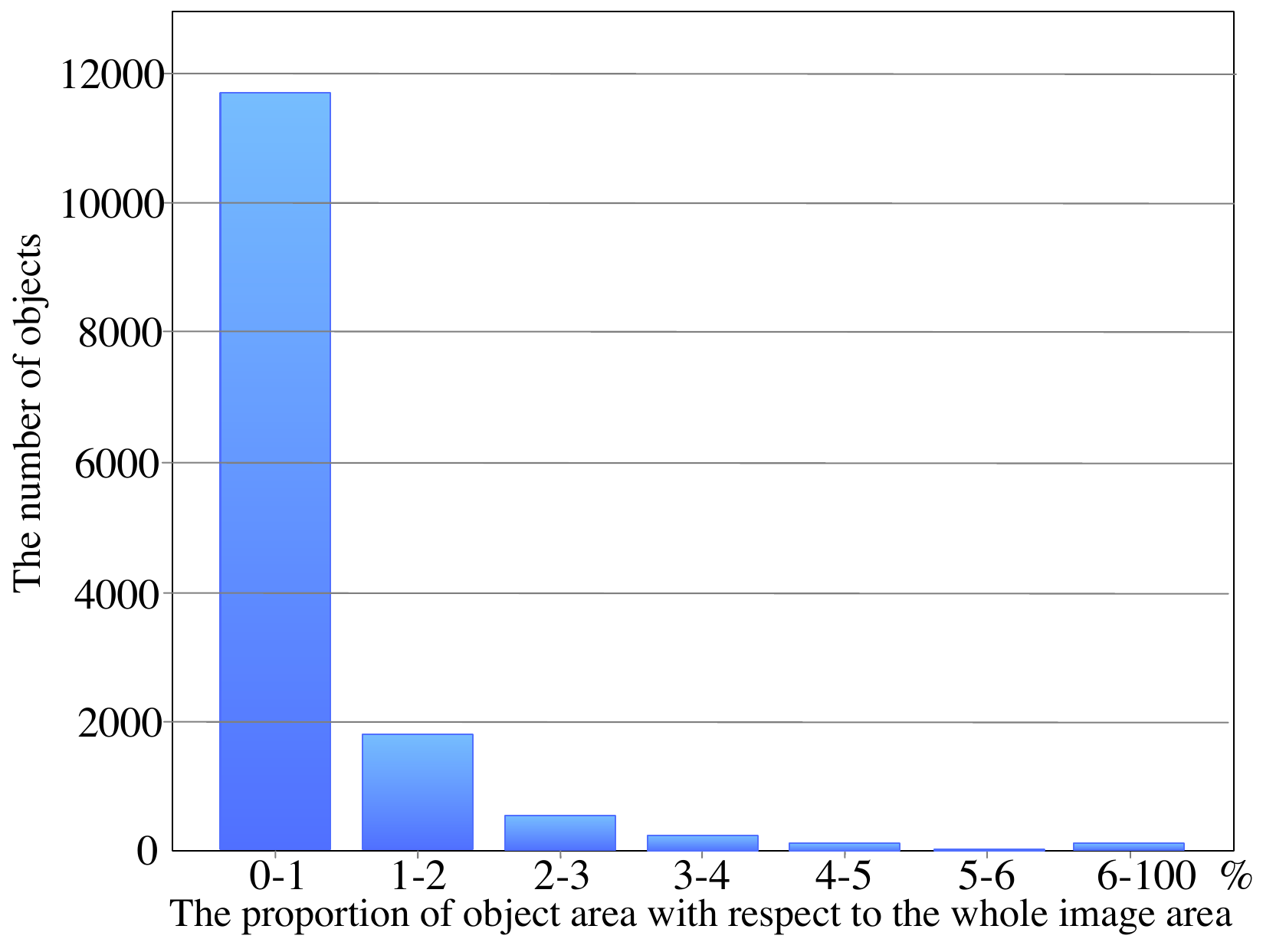}
\end{center}
\caption{The proportion distribution of the objects in our proposed UDD. It shows that the proposed dataset contains mass small objects because the breed quantities of different creatures are different within the sea farm.}
\label{scale}
\end{figure}

\subsubsection{URPC2018}It contains 2,901 images for training and 800 images for testing and the resolutions of the images are 586$\times$480, 704$\times$576, 720$\times$405, and 1,920$\times$1,080. The testing set is not available and the image quality of the whole dataset is unsatisfactory as shown in Fig. \ref {samples}. Besides, some images were also collected from an artificial underwater environment and the numbers of three creatures are 3,553, 13,609, and 1,467, which means it has the class-imbalance problem.

To overcome their drawbacks, we proposed a new benchmark UDD, which is collected in a real open-sea farm by a 4K camera carried by robots or divers.

 \begin{figure*}[!t]
\begin{center}
\includegraphics[width=0.95\linewidth]{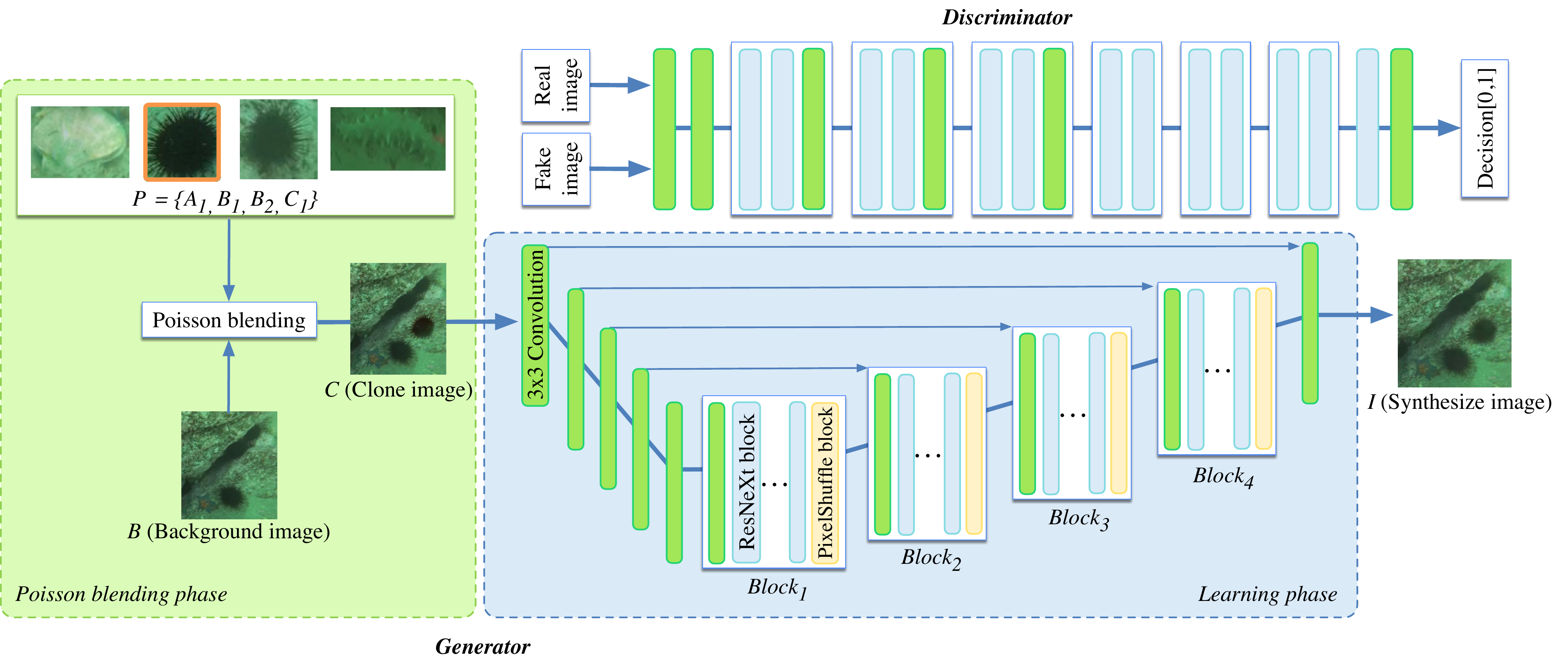}
\end{center}
   \caption{Overview of Poisson GAN. \emph{P} means a set of objects to be embedded and here we randomly choose \emph{B$_{1}$} to do Poisson blending with the background image \emph{B}.}
\label{augmentation}
\end{figure*}

\subsection{Data Augmentation with GAN} 
Data augmentation means adding more variation to the training data in order to improve the generalization capabilities of the model. Now, data augmentation strategies are widely used in training CNNs, like flipping, re-scaling, \emph{etc}. Recently, GANs \cite{Goodfellow2014Generative} have shown excellent performance on many image to image tasks \cite{9370116,9119486}. There are also some works \cite{8962219,Zhu2017Unpaired} using GAN on data augmentation. AugGAN \cite{Huang2018AugGAN} applies structure-aware semantic segmentation with soft weight-sharing, so the generated images are convincing enough to be trained. However, the ground truth used in this method contains masks, which are inconvenient to be labeled. CycleGAN \cite{Zhu2017Unpaired} only needs an unpaired dataset, thus reducing the difficulty of preparing training set, and there has been a large number of works using CycleGAN to achieve data augmentation \cite{Deng2017Image,Wei2018Person}. However, 
it may cause over-fitting easily in generating images thus affecting models' accuracy.

Existing GAN-based methods have achieved a lot in image classification by transforming the image styles, while they may not be good at object detection since they cannot change the number, size, or position of objects in an image, which is vitally important in detection. Therefore, we propose Poisson GAN to solve the problem.  
\subsection{Underwater Object Detection}
The development of underwater object detection is similar to that of general object detection. Before CNNs, detectors are mainly based on the sliding-window paradigm with hand-craft features like SIFT \cite{Lowe2004Distinctive} and HOG \cite{1467360}. Mehdi \emph{etal}. \cite{Ravanbakhsh2015Automated} adopted Haar \cite{Viola2004Robust} and shape feature in automated fish detection.

With the development of CNNs, CNN-based detectors achieve great improvement in obejct detection. Modern CNN-based object detection approaches can be roughly divided into two categories: two-stage method and one-stage method. Two-stage methods (\emph{e.g.}, Faster R-CNN \cite{Ren2015Faster}, R-FCN \cite{Dai2016R}) first propose object generation (\emph{e.g.}, EdgeBoxes \cite{Zitnick2014Edge}, RPN \cite{Girshick2015Fast}) and then determine position and class of objects. They achieve state-of-the-art performance, but need amount of computation, which cannot meet real-time requirements. One-stage methods unify the proposal and prediction processes, making detectors faster than two-stage methods. Redmon \emph{et al}. proposed YOLO \cite{Redmon2015You} using an end-to-end CNN to directly predict each object’s class and location, but there is still a large accuracy gap between YOLO and other two-stage methods. After that, SSD \cite{Liu2016SSD} adopts the concept of anchor boxes in \cite{Girshick2015Fast} and tiles anchor boxes of different scales respectively on a certain layer to improve detection performance. Recently, lots of anchor-free one-stage methods \cite{Law2018CornerNet,unknown} are proposed such as CornerNet \cite{Law2018CornerNet} and CenterNet \cite{unknown}. Inspired by the above approaches, Li \emph{et al}. \cite{Xiu2016Fast} adopt Fast R-CNN \cite{Girshick2015Fast} framework for underwater object detection. However, a detector specially designed for underwater sea farm object detection still needs to balance the speed and accuracy to meet the requirements of the robot embedding platform. Hence, we design a novel network AquaNet in this paper. 

\section{Underwater Detection Dataset}
 To provide a benchmark for researchers, we first gather an underwater dataset from a real open-sea farm and present a corresponding  comparable results. It contains 2,227 underwater images which has three types of creatures (seacucumber, seaurchin and scallop). To the best of our knowledge, this is the first 4K HD dataset collected from a real open-sea farm and is also the closest one to the real picking environment for underwater robot grabbing. In addition to harvesting sea creatures, efforts in developing these datasets would also be useful in robotic missions intended to culling sea creatures, monitoring the spatiotemporal growth of sea creatures, \emph{etc}.

\subsection{Dataset Construction}
Because the detector is a key part of the underwater picking system deployed in a robot, the best way to improve the performance of it is to use real picking environment images to train it. 
Therefore, we used an HD camera Yi 4K$\protect\footnote{Yi 4K is produced by YI TECHNOLOGY and equipped with the SONY IMX377 sensor. It supports recording native 4K/30fps, 2.7K/60fps, 1080P/120fps, and 720P/240fps videos and has been carried on many underwater robots.}$ operated by divers or robots to collect the dataset at Zhangzidao of Dalian in China on November 2018. The diver-captured images are mainly used to provide more diverse samples. 
As to the detail of the data collection, two underwater locations about 500 meters apart from Zhangzidao were selected for recording videos. We collected one robot-captured video and one diver-captured video at each location. The robot walked on the ocean floor to record 720P, 1080P, and 4K videos and the diver recorded the same resolution videos from a top view. Both of them followed a specific circular route. After recording the videos, we cut out four videos according to the different terrains (\emph{e.g.}, flat, slope, and stone) as shown in Fig. \ref {samples}. At last, we sampled images every 1,000 frames from the four videos to construct UDD.
This dataset includes 2,227 pictures where 1,827 ones are for training and 400 for testing (we choose one picture every seven in the total sequence for generating the testing set).

\subsection{Dataset Analysis}
\subsubsection{The number of instances} The total number of objects is 15,022. Seacucumber, seaurchin, and scallop are 1,148, 13,592, 282, respectively. Because of the difference of the different seafood economic benefit, the breed quantity is different. Therefore, the class-imbalance issue is very serious in UDD. 
We did not attempt to use the remedies, such as undersampling, oversampling, or SMOTE \cite{2002SMOTE}, to solve the class-imbalance problem since the samples in underwater object detection are objects, other than vectors or whole images where these kinds of methods could be worked. Object-level sample operation needs to know where we attach new objects and make them as realistic as the original one. Under this assumption, we propose Poisson GAN to balance the number of objects in different categories thus alleviating the class-imbalance problem of the collected dataset. 

\begin{table}[!t]
\caption{
Comparisons of different datasets. Res. represents the primary resolution of a dataset; Ins./image refers to the average number of instances per image; Ins. percent is the average percentage of instance to the image size.}
\centering 
\begin{tabular}{|l|c|c|c|c|}
\hline
Dataset&Res.&Ins./image&Ins. percent \%&Year\\ 
\hline  
VOC \cite{Everingham2007The}&500$\times$375&3&12.346&2012\\
\hline
COCO \cite{Belongie2014}&640$\times$480&7.7&5.672&2014\\
\hline
URPC2017&720$\times$405&9.3&0.73&2017\\
\hline
URPC2018&1920$\times$1080&8.2&0.58&2018\\
\hline
UDD (\bf ours)&\bf 3840$\times$2160 &\bf 10&\bf 0.47&2020\\
\hline  
\end{tabular}{}
\label{datatable}
\end{table}
\subsubsection{Instance size distribution} 
Table \ref{datatable} demonstrates the differences between UDD and other datasets. UDD has the highest resolution (4K) and the lowest instance percent (0.47\%). In terms of the instance number per image, UDD averages 10, while MS COCO \cite{Belongie2014} and PASCAL VOC \cite{Everingham2007The} average 7.7 and 3, respectively. In VOC or COCO, roughly 50\% of all objects occupy no more than 10\% of the image itself, and the other evenly occupy from 10\% to 100\%. However, UDD contains almost exclusively small instances as shown in Fig. \ref{scale}. Within it, more than 90\% of instances are less than 1.654\% of the image size. The average object size is only 44 $\times$ 28 pixels.
In general, smaller objects are harder to detect. Almost all previous detectors are evaluated on PASCAL VOC or MS COCO. 
However, under MS COCO's evaluation standard, the performance of detectors in detecting small objects needs to be improved a lot 
as \cite{wu2019recent} pointed out. Therefore, for a dataset containing mass small objects like our UDD, most detectors may not perform as well as they do in MS COCO or PASCAL VOC. It is necessary to design a detector to deal with mass small objects and stay high efficient at the same time for underwater robot picking. 

\section{Poisson GAN}
To solve the tough problems of insufficient data and class-imbalance in UDD, Poisson GAN with DR loss is proposed. Notably, Poisson GAN is proposed for underwater object detection while existing GAN-based augmentation methods \cite{Deng2017Image,Wei2018Person} are mainly for the classification task. 

\subsection{Network Architecture}
\subsubsection{Generator} The generator consists of Poisson blending phase and learning phase. Poisson blending phase changes the number, size, or position of objects in the input image and eliminates the boundary of the fusion part to a certain extent. Learning phase further improves the image quality to get a more realistic image through a learned mapping. 
\begin{figure}
\centering
\includegraphics[width=0.95\linewidth]{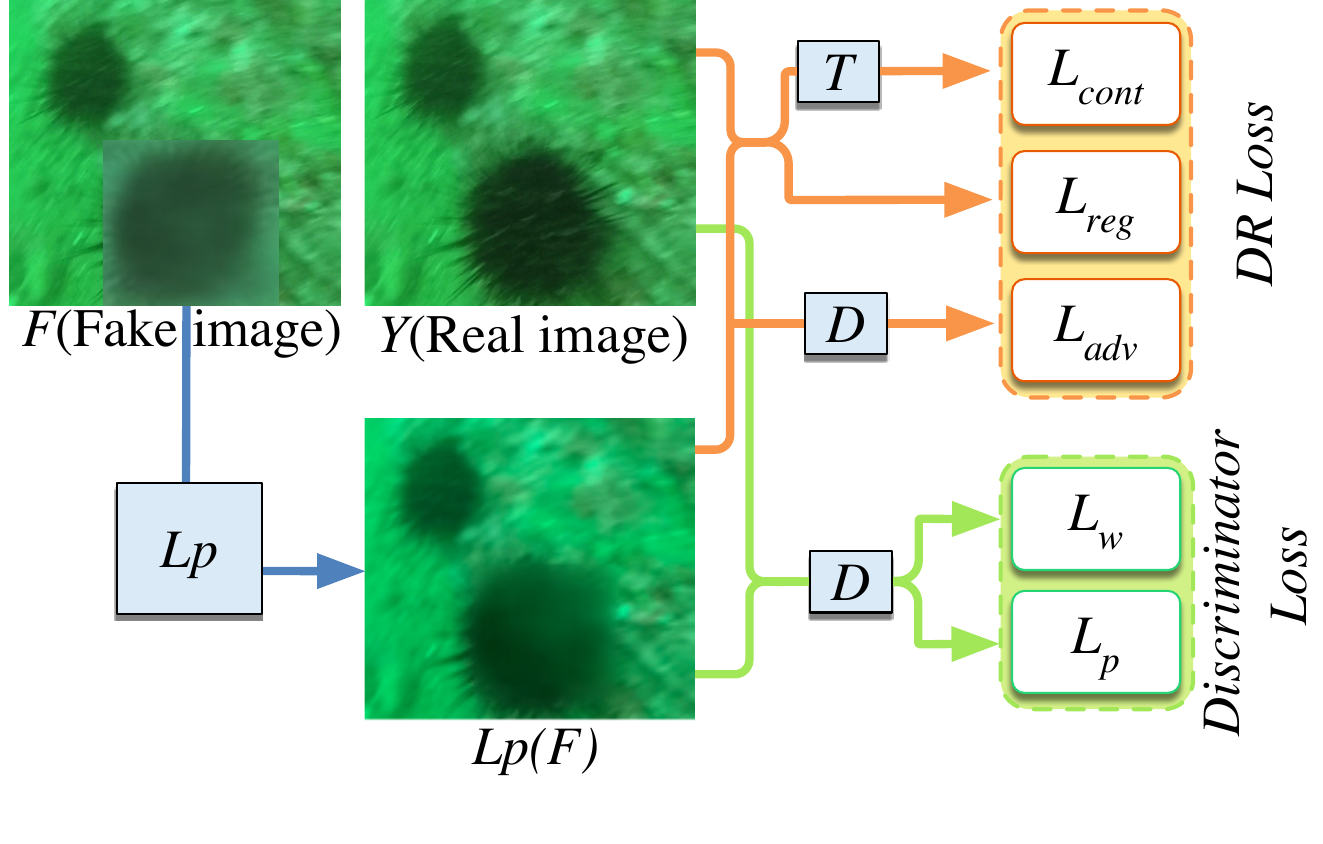}
\caption{Overview of the loss function calculation process. \emph{Lp} and \emph{D} denote the learning phase and discriminator. \emph{Lp(F)} is the output image. Extractor \emph{T} extracts semantic feature maps from \emph{Y} and \emph{Lp(F)} to calculate the \emph{L$_{cont}$}. \emph{L$_{reg}$} is calculated based on \emph{Y} and \emph{Lp(F)}, and \emph{L$_{adv}$} is calculated based on \emph{Lp(F)}. \emph{L$_{w}$} and \emph{L$_{p}$} are also calculated based on \emph{Y} and \emph{Lp(F)}. Please refer to the main body for more details.  }
\label{training}
\end{figure}
\begin{figure*}[!t]
\centering
\includegraphics[width=0.95\linewidth]{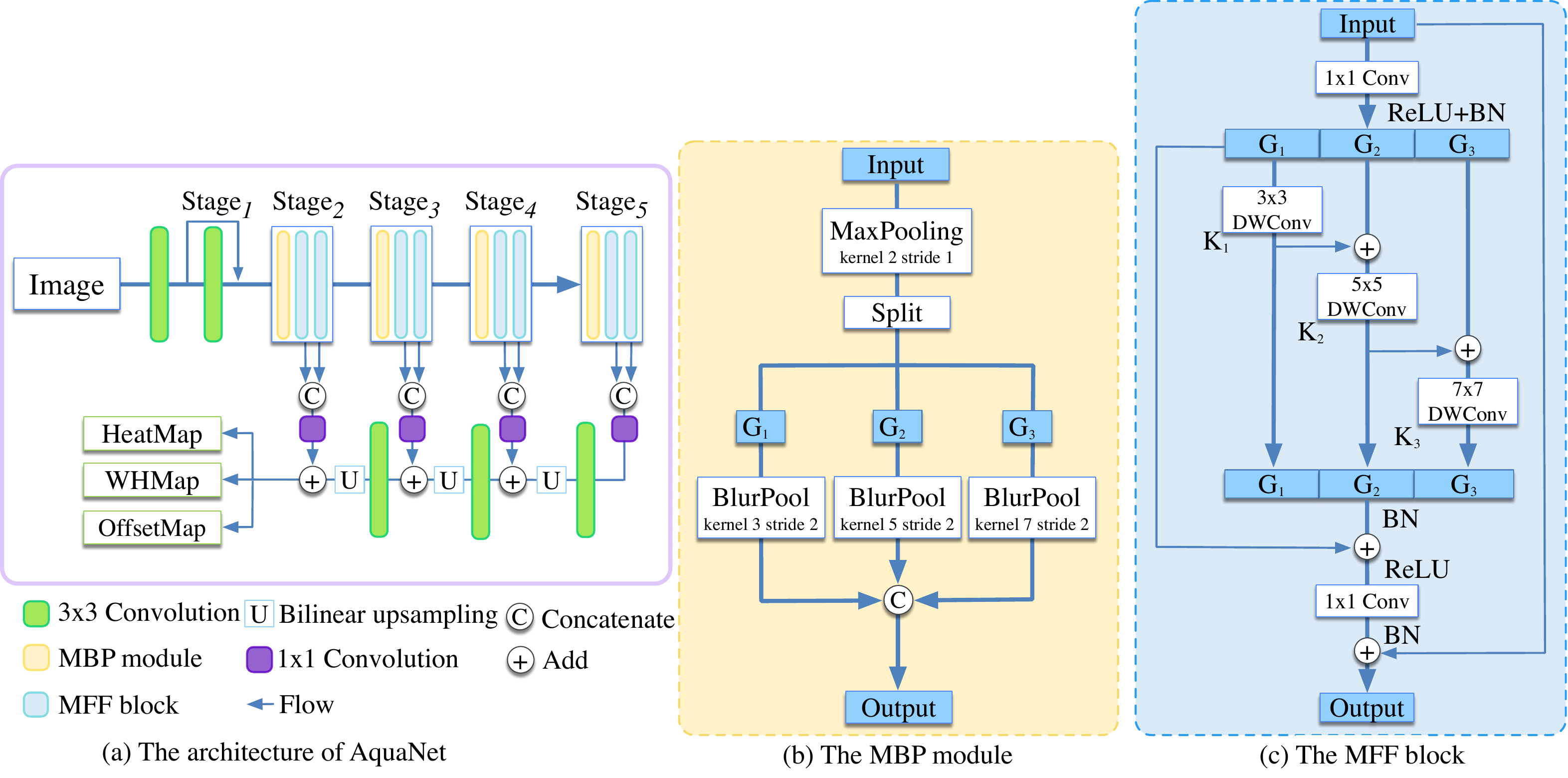}

   \caption{(a) The architecture of AquaNet. (b) The MBP module. (c) The MFF block. In (a), HeatMap predicts the center of an object; WHMap predicts the width and height of an object; OffsetMap predicts the offset between the predicted center and the real center of an object. In (b), Split means splitting the input feature map into \emph{N} (\emph{N} is the number of BlurPool \cite{Zhang2019Making} and we set it to 
  three equal groups along channel axis, and BlurPool \cite{Zhang2019Making} denotes downsampling the input feature map with a normalized Gaussian filter. In (c), DWConv denotes depth-wise convolution.}
\label{architecture}
\end{figure*}

{\bf Poisson blending phase.} Compared to the diversity of the land scenes, underwater sea farm scenes are relatively simple thus Poisson blending \cite{Rez2003Poisson} can be employed to create relatively real pictures. We embed Poisson blending into generator to change the number, position, or size of objects when generating AUDD. The process of Poisson blending is illustrated in Fig. \ref{augmentation} left. We first crop \emph{X}, \emph{Y}, and \emph{Z} seacucumbers, seaurchins and scallops from UDD respectively to build an object set \emph{P}:
\begin{equation}
P = \{A_{1},A_{2}...A_{X},B_{1},B_{2}...B_{Y},C_{1},C_{2}...C_{Z}\},
\end{equation}
where \emph{A}, \emph{B}, \emph{C} indicate seacucumber, seaurchin, and scallop, respectively. For each generation, we randomly select objects from \emph{P} to do Poisson blending with a background image \emph{B} $\in$ \emph{$\mathbb{R}$$^{3 \times 720 \times 405}$} to get a clone image \emph{C} $\in$ \emph{$\mathbb{R}$$^{3 \times 720 \times 405}$}. Notably, we try to put the embedded objects in the vicinity of the same categories in order to ensure the rationality of the clone images. Because Poisson blending works perfectly only when the contours of embedded objects are well labeled, we also need the learning phase to get more realistic images.

{\bf Learning phase.} Learning phase is inspired by \cite{Ci2018User}, an anime line art colorization GAN. We use U-net \cite{Ronneberger2015U} as backbone structure as shown in lower right part of Fig. \ref{augmentation} and a stack of 3$\times$3 convolution layers is utilized to construct the encoder. For the decoder, 4 stacks of ResNeXt blocks \cite{Xie2016Aggregated} are employed, 
denoted as \emph{Block$_{n}$}, \emph{n} $\in$ \{1,...,4\}. In the experiments, we set \emph{Block$_{n}$} to [20,10,10,5].

\subsubsection{Discriminator} The discriminator is also built with stacks of ResNeXt blocks and convolution layers as shown in upper  right part of Fig. \ref{augmentation}. The architecture is similar to the setup of SRGAN \cite{Christian2017Photo} with some modifications. We employ the same basic block used in the generator. We additionally stack more layers so that it can process 512$\times$512 inputs.
\subsection{Image Pairs Used to Train Poisson GAN} 
To further eliminate the boundaries produced by Poisson blending phase, the learning phase needs to learn a boundary elimination map to create a more realistic image. Therefore, image pairs should only differ in the edge information of the fusion parts. As we discussed in Poisson blending phase, underwater sea farm scenes are relatively simple. Hence, we can create the fake image by pasting the objects cropped from clone images \emph{C} to the places of the same category objects in the real image directly. In view of appearance similarity of the same species, 2 images (the real image and the fake image) within one image pair are nearly the same except the edge information.
\subsection{Loss Function} In general, the loss function in GAN contains a generator loss and a discriminator loss. The whole loss function calculation process is shown in Fig. \ref{training}. By leveraging the annotation information, we design DR loss as the generator loss. With a combination of content loss and region loss, DR loss can supervise the GAN both on pixel level and feature level to generate more realistic outputs. The DR loss ${G}$ is defined as:
\begin{equation}
{L}_{{G}}={L}_{cont}+{L}_{reg}+\lambda_{1} {L}_{adv},
\end{equation}
where $\lambda_{1}$ equals to 1$\times10^{-4}$ in all experiments. The content loss \emph{${L}_{cont}$} is used to supervise the model in the implicit space and is defined as:
\begin{equation}
{L}_{c o n t}=\frac{1}{c\cdot h\cdot w}\left\|T\left({Lp}\left(F\right)\right)- T(Y)\right\|_{2}^{2},
\end{equation}
in which \emph{c}, \emph{h}, \emph{w} denotes the number of channels, height and width of the output feature map. \emph{T()} denotes the feature maps obtained by the 4th convolution layer (after activation) within the VGG16 network pretrained on ImageNet. \emph{F} and \emph{Y} are the fake image and the real image within one image pair. \emph{Lp(F)} means the output image.  

The region loss \emph{${L}_{reg}$} can supervise the model in the image space and is defined as:
\begin{equation}
{L}_{r e g}=\frac{1}{c\cdot h\cdot w}\left\|\left({Lp}\left(F \right)-Y\right) \cdot M\right\|,
\end{equation}
where \emph{M} $\in$ \emph{$\mathbb{R}$$^{1 \times w \times h}$} denotes the mask created by bounding boxes of embedded objects. We take the value of 100 for the pixels in boxes while the value of other areas is 0.1. 

Adversarial loss \emph{${L}_{adv}$} is used to distinguish between real and fake training pairs and can be defined as:
\begin{equation}
{L}_{a d v}=-\mathbb{E}_{{Lp}\left(F) \sim P_{{Lp}}\right.}\left[{D}\left({Lp}\left(F\right)\right)\right].
\end{equation}

Discriminator loss are formulated as a combination of wasserstein critic loss and penalty loss:
\begin{equation}
{L}_{{D}}={L}_{w}+{L}_{p}.
\end{equation}

While the critic loss is the WGAN \cite{arjovsky2017wasserstein}:
\begin{equation}
{L}_{w}=\mathbb{E}_{{Lp}\left(F\right) \sim P_{{Lp}}}\left[{D}\left({Lp}\left(F\right)\right)\right]-\mathbb{E}_{Y \sim P_{r}}[{D}\left(Y\right)].
\end{equation}
For the penalty term, we combine the gradient penalty \cite{gulrajani2017improved} and an extra constraint term introduced by Karras \emph{et al}. \cite{karras2017progressive}:
\begin{equation}
\begin{split}
{L}_{p}&=\lambda_{2} \mathbb{E}_{\widehat{Y} \sim P_{i}}\left[\left(\left\|\nabla_{\hat{Y}} {D}(\hat{Y})\right\|_{2}-1\right)^{2}\right]\\
+&\epsilon_{d r i f t} \mathbb{E}_{Y \sim P_{r}}\left[{D}(Y)^{2}\right],
\end{split}
\end{equation}
With the penalty loss, the output value can keep stable near zero and the training time can be reduced greatly. In the experiment, we set $\lambda_{2} = 10$ and $\epsilon_{d r i f t} = 1 \times 10^{-3}$. Besides, the distribution of interpolated points \emph{$P_{i}$} is defined as following:
\begin{equation}
\hat{Y}=\epsilon {Lp}\left(F\right)+(1-\epsilon) Y , \epsilon \in U[0,1].
\end{equation}
As such, we penalize the gradient over straight lines between points in the real distribution \emph{$P_{r}$} and generator distribution \emph{$P_{{G}}$}.

\subsection{Expanded Dataset}
 To build the object set \emph{P}, we crop 1,000 seaurchins, 150 seacucumbers,  and 35 scallops from UDD, and then fuse them into background images through Poisson GAN.

\subsubsection{AUDD dataset} We use images from UDD as background images to generate synthesize images via Poisson GAN. Each image in UDD training set is copied into 11 images and one of them is added into AUDD as the original one. Other 10 images take paste operation 0, 1, 2, 3 times with a probability of 0.1, 0.35, 0.35, 0.2 and in one paste operation, Poisson blending is performed 0, 1, 2 times with a probability of 0.3, 0.5, 0.2. Finally, only processed images will be added to AUDD. As a result, this dataset contains 18,661 images and has 18,350 seacucumbers, 101,422 seaurchins, and 9,624 scallops. These images are nearly realistic and can be seen as a supplement of UDD as shown in Fig. \ref{comparison3}. 

\subsubsection{Pre-training dataset} We put cropped instances and background images from URPC (URPC2017 and URPC2018) or other underwater pictures that we captured into the Poisson blending phase to build the pre-training dataset. The main purpose of this dataset is to provide a more robust pre-trained model for a detector. Images in it are divided into three classes (hard, normal, and easy) according to the number or size of objects in an image. The hard one contains 8-10 objects whose shape is correspondingly tiny while the normal one includes 4-8 medium-sized objects. Images in the third one contain only 2-3 large objects. Each class contains about 197,000 images (195,400, 198,700, 191,980, respectively). Therefore, there are up to 589,080 images including different background colors, points of view, and terrains to make sure detectors achieve a good generalization. 

\subsection{Discussion}
Poisson GAN can change the background of an instance while preserving the properties of the instance. Therefore, besides augmenting the underwater images for training detectors, Poisson GAN may be applied to other works such as Pseudo-label-based unsurprised domain adaptation within Unsupervised Person re-identification. In the Pseudo-label-based methods, Poisson GAN (trained in the source domain or the joint of the source and target domain) may augment the clustered
pseudo-label images to mitigate the effects of pseudo label noise and enrich the target domain data.

\section{AquaNet}
As we discussed at the end of Section \uppercase\expandafter{\romannumeral3}, we need to design a detector to efficiently detect a large number of small objects in cloudy underwater images.
Therefore, AquaNet with two simple yet efficient modules are designed. 

\subsection{Multi-scale Blursampling}
Detecting small objects from cloudy  underwater images requires the high robustness of a detector. However, as BlurPool \cite{Zhang2019Making} points out, normal downsampling methods (MaxPooling, AveragePooling, and the convolution with stride 2) do not have the capability to anti-alias, which could reduce the robustness of CNNs. To address this problem, we design the MBP module inspired by MixNet \cite{1907.09595} and BlurPool. It can generate diverse downsampling feature maps and add only a little computation burden compared with MaxPooling.
\begin{figure}[!t]
\begin{center}
\includegraphics[width=1\linewidth]{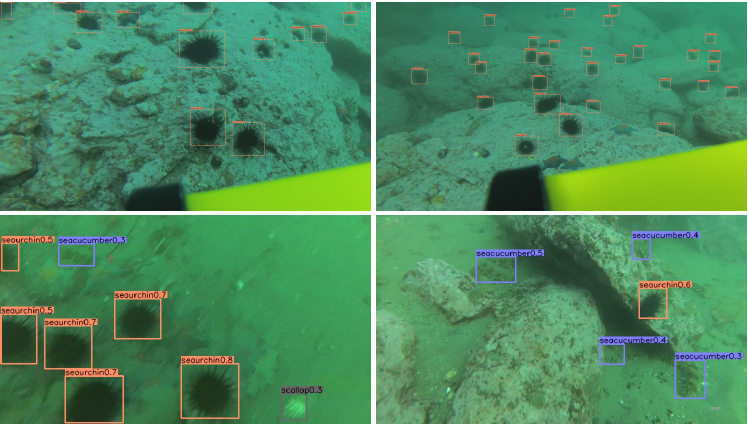}
\end{center}
   \caption{Visualization results of AquaNet. The results show that detecting mass small objects and solving class-imbalance problem are important to underwater object detection for robot picking.
}
\label{samples1}
\end{figure}
Fig. \ref{architecture} (b) shows the structure of MBP module, which is composed of one MaxPooling with stride 1 and several BlurPool with stride 2. For an input feature map, it is first processed by the MaxPooling and split into \emph{N} (\emph{N} is the number of BlurPool) groups along channel axis. Then, each group is processed by an independent BlurPool. Finally, we concatenate all the groups to get the final output. In all the experiments, we set \emph{N} to 3.

BlurPool is an operation using a normalized Gaussian filter with stride 2 to downsample an input feature map. The method allows for a choice of blur kernel. In our module, we choose the following three different sizes' filters ranging with increasing smoothing: {\bf Triangle-3} [1, 2, 1], {\bf Binomial-5} [1, 4, 6, 4, 1], {\bf Binomial-7} [1, 6, 15, 20, 15, 6, 1]. The weights are normalized, and the filters are the outer product of the following vectors with themselves.
\subsection{Multi-scale Contextual Features Fusion}
To get a better performance in detecting small objects, a backbone needs to extract multi-scale features as much as possible. Many approaches \cite{Chen2018DeepLab,Li} choose to fuse features after backbone extracts features, while this means more layers are needed. Others \cite{Gao2019Res2Net,1907.09595} choose the way using different sizes of kernels or complex connection in one block. 
Inspired by the latter methods and considering the efficiency requirements in our application field, we combine depth-wise convolution and the connection method in Res2Net \cite{Gao2019Res2Net} to create the MFF block, which can extract and fuse multi-scale features at the same time.
\begin{table*}[!t]
\caption{Comparisons on UDD testing set. Frame-per-second (FPS) denotes the reciprocal of the sum of inference and post-processing time measured on the same machine. mAP$_{cu}$, mAP$_{ur}$ and, mAP$_{sc}$ denote the mAP$_{50}$ of seacucumber, seaurchin, and scallop, respectively. \emph {1x} denotes scaling the corresponding baseline models with depth multiplier 1.0. MNet means the backbone of AquaNet in Fig \ref{architecture} (a).}
\centering 
\begin{tabular}{|l|c|c|cc|c|c|c|c|}
\hline
	\multirow{2}{*}{Method}&\multirow{2}{*}{Backbone}&\multirow{2}{*}{Params}&\multicolumn{2}{c|}{FPS}&mAP$_{50}$&mAP$_{cu}$&mAP$_{ur}$&mAP$_{sc}$\cr\cline{4-9}

	&&&TITAN&XAVIER&\multicolumn{4}{c|}{scratch/pre-trained} \\
\hline
YOLOv3 \cite{Redmon2018YOLOv3}&DarkNet-53&61.9M&32&2.1/4.9/5.0& 46.8/{\bf57.8}&30.7/52.6&77.5/80.3&32.3/40.6 \\
\hline
SSD \cite{Liu2016SSD}&MobileNetV2&3.05M&11&4.1/5.5/6.0&22.7/38.9&14.6/18.9 &48.1/75.3 &5.3/22.4 \\
\hline
RetinaNet \cite{Lin2017Focal}&ResNet-18&19.81M&14&3.9/6.2/6.2&24.6/49.7&3.1/50.7&61.3/70.3&9.4/28.1 \\
\hline
RetinaNet \cite{Lin2017Focal}&ResNet-50&36.15M&10&2.7/5.0/5.3&34.2/50.6&15.2/51.1&65.1/71.7&22.2/28.9 \\
\hline
FCOS \cite{DBLP:journals/corr/abs-1904-01355}&ResNet-50&31.84M&27&3.4/6.6/6.7 & 44.9/55.3& {\bf 35.5}/50.4 & 73.9/79.4 &25.3/36.1 \\
\hline
Foveabox \cite{DBLP:journals/corr/abs-1904-03797}&ResNet-50&36.02M&28&3.2/6.8/7.9 & 30.0/54.1 & 16.1/50.7 & 61.4/76.8 &12.6/34.7 \\
\hline
FreeAnchor \cite{zhang2019freeanchor}&ResNet-50&36.15M&25&2.8/5.0/5.4 & 32.7/56.9 & 17.3/{\bf54.9} & 71.0/81.0 &9.8/34.8 \\
\hline
RPDet \cite{DBLP:journals/corr/abs-1904-11490}&ResNet-50&36.6M&22&2.2/3.4/3.4 & 45.1/57.7 & 26.9/54.1 & 76.1/{\bf83.9}&32.4/35.0 \\
\hline
GA-RetinaNet \cite{wang2019region}&ResNet-50&37.15M& 12&2.1/3.4/3.4& 36.1/48.2 & 21.7/40.5& 70.5/79.9 &16.1/24.3\\
\hline
CenterNet \cite{unknown}&DLA-34 \cite{Yu2017DeepLA}&18.12M&33&4.0/7.7/9.8 & 36.6/46.0& 12.5/32.7 & 78.0/78.3 &19.4/27.1 \\
\hline  
AquaNet&MobileNetV3-small (1x) \cite{Howard2019SearchingFM}& 1.92M& 45&5.3/7.8/9.8&39.1/40.6&15.4/17.4& 77.6/77.7& 24.5/26.6 \\
\hline
AquaNet&ShuffleNetV2 (1x) \cite{Ma2018ShuffleNetVP}& 1.88M& 32&5.4/8.0/{\bf10.0}&37.7/43.0&11.4/23.9& 74.9/77.5& 27.0/27.6 \\
\hline
AquaNet&EfficientNet-B0 \cite{Tan2019EfficientNetRM}& 4.59M& 44&5.0/7.2/8.5&34.7/40.3&9.9/22.4& 77.5/78.8& 16.8/19.8 \\
\hline
AquaNet&MixNet-S (1x) \cite{1907.09595}& 3.27M& 23&4.3/6.0/6.9&36.2/43.3&9.3/22.1&76.6/77.0&22.6/30.8\\
\hline
AquaNet&MnasNet-A1 (1x) \cite{Tan2018MnasNetPN}& 3.18M& 29&{\bf5.6}/8.5/9.7&37.0/43.4&11.5/24.6& 77.8/80.0& 21.8/25.6 \\
\hline  
AquaNet (\bf ours)&MNet&\bf 1.30M&\bf 48&{\bf5.6}/{\bf9.1}/{\bf10.0}&{\bf47.4}/55.3&23.6/38.9&{\bf 79.3}/80.1&{\bf 39.4}/{\bf47.0} \\
\hline
\end{tabular} 

\label{t3}
\end{table*}
Fig. \ref{architecture} (c) shows the structure of MFF block. For an input, the number of channels is first expanded \emph{N} times (\emph{N} is the number of kernel sequence we set, \emph{e.g.}, in Fig. \ref{architecture} (c) the sequence is [3, 5, 7]) by  1 $ \times$ 1 convolution. Similar to the Res2Net module, the output feature map is split into \emph{N} groups by channel, denoted by \emph{G$_{i}$}, \emph{i} $\in$ \{1,...,\emph{N}\}. Then, each group is processed by a depth-wise filter \emph{K$_{i}$} with the corresponding kernel size in kernel sequence. The output of \emph{K$_{i}$} is added to the following subset \emph{G$_{i+1}$}, and then processed by \emph{K$_{i+1}$}. The outputs of these parallel branches are concatenated and then fused by the final 1 $ \times$ 1 convolution to reduce output channels. Besides, we use two residual connections in our block: one is between the input and output feature map, the other is between the expanded feature maps. The final output of MFF block is a feature map aggregating multi-scale features, which is vitally important in detecting small objects in cloudy underwater images.

\subsection{Network Architecture}
The entire architecture is shown in Fig. \ref{architecture} (a). Because of the characteristics of being anchor-free and NMS-free, we adapt CenterNet \cite{unknown} for our detector to improve the detection efficiency. Different from previous hand-crafted backbone or the one searched from ImageNet, only two 3$\times$3 convolution layers and 8 MFF blocks with MBP are used to build our backbone because the features of small objects may disappear or be distorted if there are too many layers in a backbone. The kernel sequences (see MFF) from Stage$_{2}$ to Stage$_{4}$ are set into [3, 5, 7], while in Stage$_{5}$ they are [3, 5, 7, 9]. To enhance the inner information flow, we use the concatenation of the output of the last 2 layers of Stage$_{2,3,4,5}$ to fuse when upsampling. Finally, we adopt the same detection head algorithm as CenterNet defines. The head has 3 maps: HeatMap that predicts the center of an object; WHMap that estimates the width and height of an object; OffsetMap that predicts the offset between the predicted center and the real center of an object. For each map, the features of the backbone are passed through a 3$\times$3 convolution, ReLU, and another 1$\times$1 convolution layer. The resolution is a quarter of the input resolution. The loss function is also as same as that in \cite{unknown}. In object detection community, a detector can only predict 0 for a pixel belonging to the out-of-distribution (OOD) samples. Our detector also employs the same mechanism.

\subsection{Discussion}
Some works \cite{9321130,9055436,8762119} also explore various RNN-like pooling/fusing strategies. The various RNN or LSTM structures play an important role in understanding contextual information, and the idea of RNN has also inspired the design of MFF. However, the main difference between AquaNet and aforementioned works is that AquaNet should achieve the purpose of detecting a large number of small objects from cloudy underwater images on the premise of giving consideration to efficiency. Therefore, we propose MFF and MBP under the guidance of the efficiency and accuracy of the underwater object detection network. 
The lightweight structure of MFF and MBP can effectively and quickly integrate multi-scale features to achieve fast small object detection, while others rarely take into account both efficiency and accuracy.
\begin{figure*}[!t]
\centering
\includegraphics[width=1\linewidth]{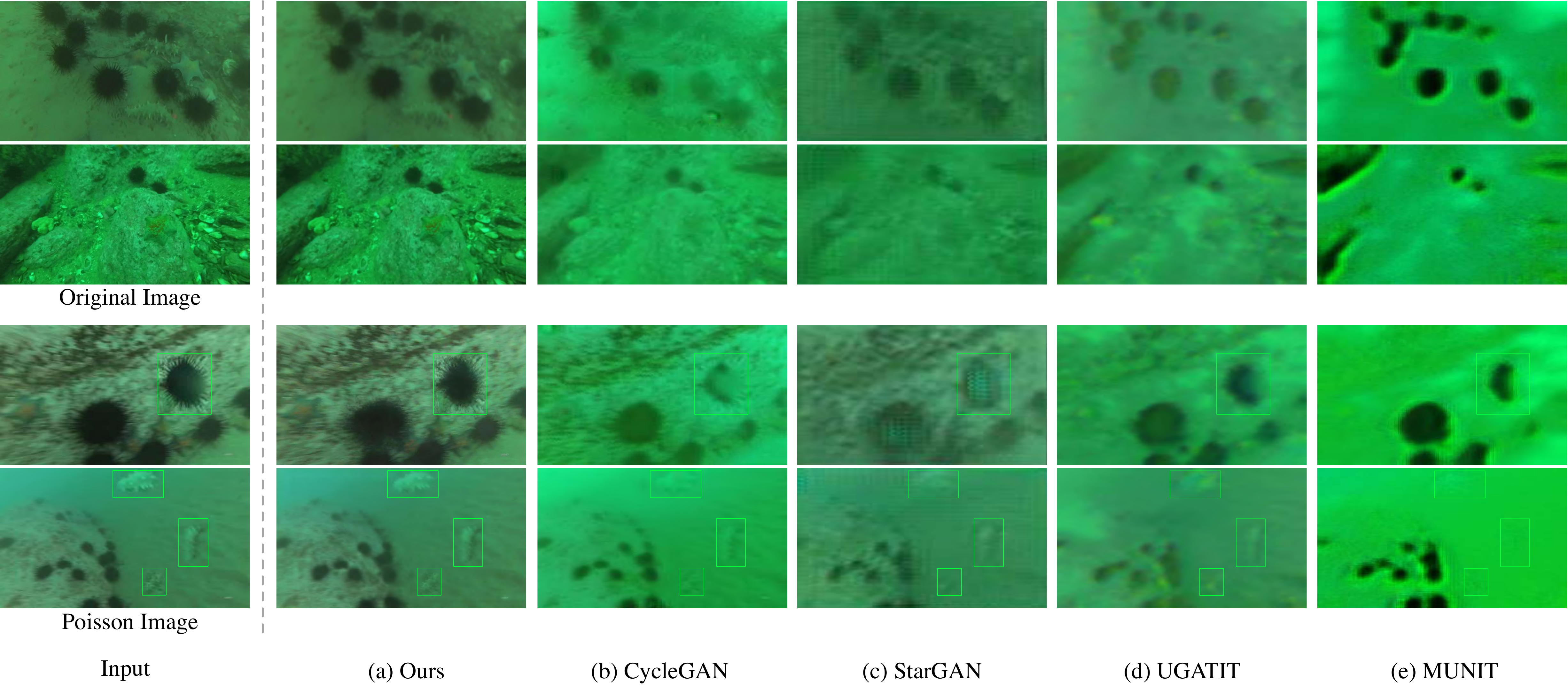}

   \caption{Comparisons of different GANs. (a) Ours, (b) CycleGAN \cite{2017arXiv170310593Z}, (c) StarGAN \cite{DBLP:journals/corr/abs-1711-09020}, (d) UGATIT \cite{DBLP:journals/corr/abs-1907-10830}, (e) MUNIT \cite{huang2018munit}. Images generated by other methods may change the texture of objects or even disappear some objects. In addition, the class-imbalance problem cannot be solved by the transformation.}
\label{comparison_2}
\end{figure*}

\section{Experiments}
\subsection{AquaNet}
\subsubsection{Experimental Settings} All the experiments are conducted with CUDA 10.0 and cuDNN 7.3.1 backends on NVIDIA TITAN XP GPUs, Intel Xeon CPU E5-2,680 v4. Our AquaNet is implemented on PyTorch. The input image size is 512$\times$512 both in training and inference. Lookahead optimizer \cite{Zhang2019Lookahead} with Adam \cite{Kingma2014AdamAM} is employed and initial learning rate is set to 2.3$\times$10$^{-5}$. Batch size is 32 and the detector is trained on one GPU. We employ the zero-mean normalization, random flip, random scaling (between 0.6 to 1.3), and cropping for data augmentation.

\begin{table}[!t]
\caption{Comparisons of different pooling strategies. \emph{k} means the kernel size in MaxBlurPool.}
\centering 
\begin{tabular}{|l|c|}
\hline
Pool&mAP$_{50}$ \\ 
\hline  
MaxPooling&41.9  \\
\hline
MaxBlurPool (\emph{k}=3) \cite{Zhang2019Making}&44.9\\
\hline
MaxBlurPool (\emph{k}=5) \cite{Zhang2019Making}&46.0\\
\hline
MaxBlurPool (\emph{k}=7) \cite{Zhang2019Making}&46.4\\
\hline
MBP (ours)&\bf 47.4\\
\hline  
\end{tabular}{}
\label{t2}
\end{table}

\begin{table}[!t]
\caption{Results of MFF block with different settings. Different kernel denotes the kernel sequence is [3, 5, 7] or [3, 5, 7, 9] if there is a checkmark, otherwise the sequence is [3, 3, 3] or [3, 3, 3, 3]. Skip connection means using the connections between subbranches and between expanded feature maps if there is a checkmark.}
\centering 
\begin{tabular}{|c|c|c|}
\hline
Different kernel&Skip connection&mAP$_{50}$ \\ 
\hline  
  &  &42.8  \\
  \hline
\checkmark & &43.1\\
\hline
\checkmark &\checkmark &\bf 47.4\\
\hline
\end{tabular}
\label{t1}
\end{table}

\subsubsection{Ablation Studies} We evaluate AquaNet on UDD testing set to investigate the effect of each part of it. All models were trained from scratch for 1600 epochs.

{\bf Ablation on MBP.} MaxPooling and MaxBlurPool \cite{Zhang2019Making} with different kernel sizes are used to compare with MBP as shown in Table \ref{t2}. Our method is 5.5\% higher than MaxPooling due to anti-aliasing and multi-scale. Our method also achieves the highest accuracy among MaxBlurPools with a single kernel size, indicating that the multi-level blur strategy is better for small object detection in cloudy underwater images.

{\bf Ablation on MFF.} Compared with the original block in MobileNetv2 \cite{Sandler2018MobileNetV2}, we apply different kernel sizes and add skip connections between branches in MFF. Table \ref{t1} shows results with different settings. For the MFF, the first row equals to the block in MobileNetV2 \cite{Sandler2018MobileNetV2}, the second one equals to the block in MixNet \cite{1907.09595}, and the third one is a standard MFF block. As we can see, different kernel and skip connection operations improve the accuracy by 4.6\% compared to the first one. The results demonstrate that our MFF block is more suitable than blocks in MobileNetv2 or MixNet to extract and represent features for the underwater open-sea scenes under the same layer configurations.
 
\subsubsection{Comparison With Other Methods on UDD} We take some public real-time methods from mmdetection$\protect\footnote{https://github.com/open-mmlab/mmdetection}$ \cite{chen2019mmdetection} and lightweight backbones to compare with AquaNet. Quantitative results are recorded in Table \ref{t3} together with some detection results shown in Fig. \ref{samples1}. 

{\bf Implement details.}
All the detectors are trained on UDD training set with scratch or pre-trained models from pre-training dataset. 
Besides, SyncBN \cite{Chao2018MegDet} is used to keep the mini-batch size equal to 32 if necessary. For detectors from mmdetection, SGD with warmup is employed. For AquaNet with different backbones, the settings are the same as that in Section \uppercase\expandafter{\romannumeral6} \emph{A 1)}. We do not employ any test augmentation (\emph{e.g.}, flip test or multi-scale test) while employ the same data augmentation mentioned (zero-mean normalization, random flip, random scaling, and cropping). About speed study, we employ two computing platforms, \emph{i.e.}, NVIDIA TITAN XP and JETSON AGX XAVIER$\protect\footnote{JETSON AGX XAVIER is an embedded development board produced by NVIDIA. Please refer https://developer.nvidia.com/embedded/jetson-agx-xavier-developer-kit for more information.}$, to study the speed and define FPS as the reciprocal of the sum of inference and post-processing time. We also take three different power modes (mode1\_10W, mode2\_15W, and mode3\_30W) on XAVIER to evaluate the detectors' speed.   

{\bf Accuracy study.}
In Table \ref{t3}, AquaNet achieves the highest mAP$_{50}$ on model trained from scratch and gets a competitive performance (55.3\% mAP$_{50}$) on pre-trained model.
Among all the methods, AquaNet could give a convincing detection result with the minimum parameters (only 1.3 million). About the two different training initializations, the loss curve of AquaNet is shown in Fig. \ref{scale} where the pre-trained curve converges faster than the scratch curve. Although the two final loss numbers are almost the same, the performances are totally different (mAP: 47.4 vs. 55.3). The more parameters within a detector, the higher chance it will overfit. When training from scratch, our detector achieves the best performance (some of the other methods may overfit considering the low performance and the large parameter number), showing that ours has a great capability of anti-overfitting. 

\begin{figure}[h]
\begin{center}
\includegraphics[width=1\linewidth]{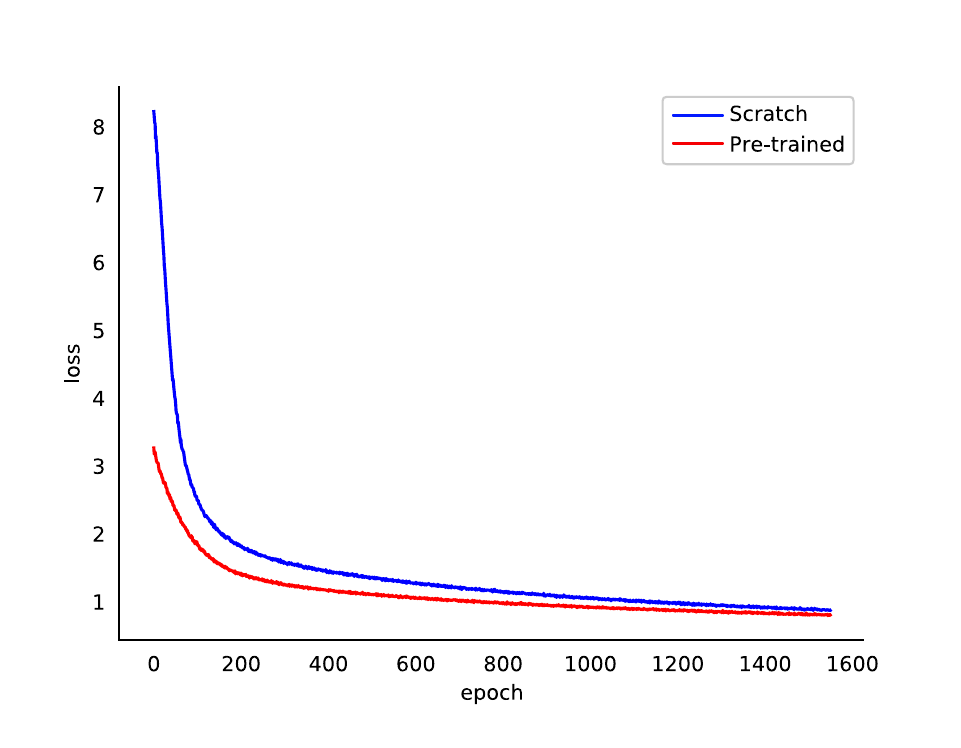}
\end{center}
\caption{The training loss of AquaNet with different initializations. The red curve means the loss trained from scratch and the blue one is trained from our pre-trained model.}
\label{scale}
\end{figure}

{\bf Speed study.}
Table \ref{t3} has shown the efficiency evaluation on two computing platforms, \emph{i.e.}, JETSON AGX XAVIER and NVIDIA TITAN XP. On JETSON AGX XAVIER, we test all methods under three power modes, \emph{i.e.}, mode1$\_$10W, mode2$\_$15W, and mode3$\_$30W, respectively; AquaNet achieves the highest speed (5.6/9.1/10.0 FPS) among all the three modes. 
On TITAN XP, AquaNet is also the fastest method with 48 FPS.
Compared with ShuffleNet V2, although depthwise separable convolution (DWS) is not as efficient as it claimed in theory, DWS with large kernel size (5 or 7) could get the same receptive field and significantly reduce the computation compared with the normal convolutional layer under the same kernel size. Besides, DWS is not the only factor that influences the efficiency. The stack number of convolutional layers, channel numbers, or even the running platform also influences the network speed. For the speed comparison between ShuffleNet V2 and our network, we just use 8 MFF and 4 MBP in our backbone, while ShuffleNet V2 uses 16 blocks (including the downsampling block). In addition, the input channels in our backbone are [32, 40, 80, 120] from Stage 2 to 5, which are narrower than that in ShuffleNet V2 ([24, 116, 232, 464]). As a result, the speed of ours surpasses ShuffleNet V2 on TITAN in Table \ref{t3}. However, the speed of them are almost the same on XAVIER, which means there are still other factors affecting the speed.

The above studies indicate that AquaNet with MBP and MFF can deal with the characteristics of UDD and finally achieve an inspiring performance and speed. However, all the methods fail to retain a higher performance on detecting seacucumber and scallop because of the insufficient training of minority classes. Therefore, AUDD is further constructed by Poisson GAN to address this issue. 

\subsection{Poisson GAN}
\subsubsection{Experimental Settings} 
Our Poisson GAN is also implemented on PyTorch. The input size is 512$\times$512 for both training and inference. Adam is employed and the learning rate is initialized with 1$\times 10^{-4}$ in both generator and discriminator, then decreased to 1$\times 10^{-5}$ after 125,000 iterations. Our experiments are conducted on a single NVIDIA TITAN XP GPU and the batch size is 4. We resize the image pairs with shorter side to be 512 and then randomly crop to 512$\times$512 before random horizontal flipping. During the training of Poisson GAN, we generated 20,000 image pairs and employed augmentation methods (\emph{i.e.}, randomly crop and random horizontal flipping) to avoid the mode collapse. 
\begin{table}[!t]
\caption{Results of training YOLOv3 on AUDD with different initializations.}
\centering 
\begin{tabular}{|l|c|c|c|c|}
\hline
Initialization&mAP$_{50}$&mAP$_{cu}$&mAP$_{ur}$&mAP$_{sc}$\\ 
\hline  
Scratch & 68.8 & 55.8& 81.4 & 69.2 \\
\hline 
ImageNet & 71.5 & 72.9 & 71.2 & 70.5 \\ 
\hline
Ours & \bf 80.5&\bf76.7&\bf84.2&\bf80.7 \\ 
\hline  
\end{tabular}
\label{t6}
\end{table}

\begin{figure*}[!t]
\begin{center}
\includegraphics[width=1\linewidth]{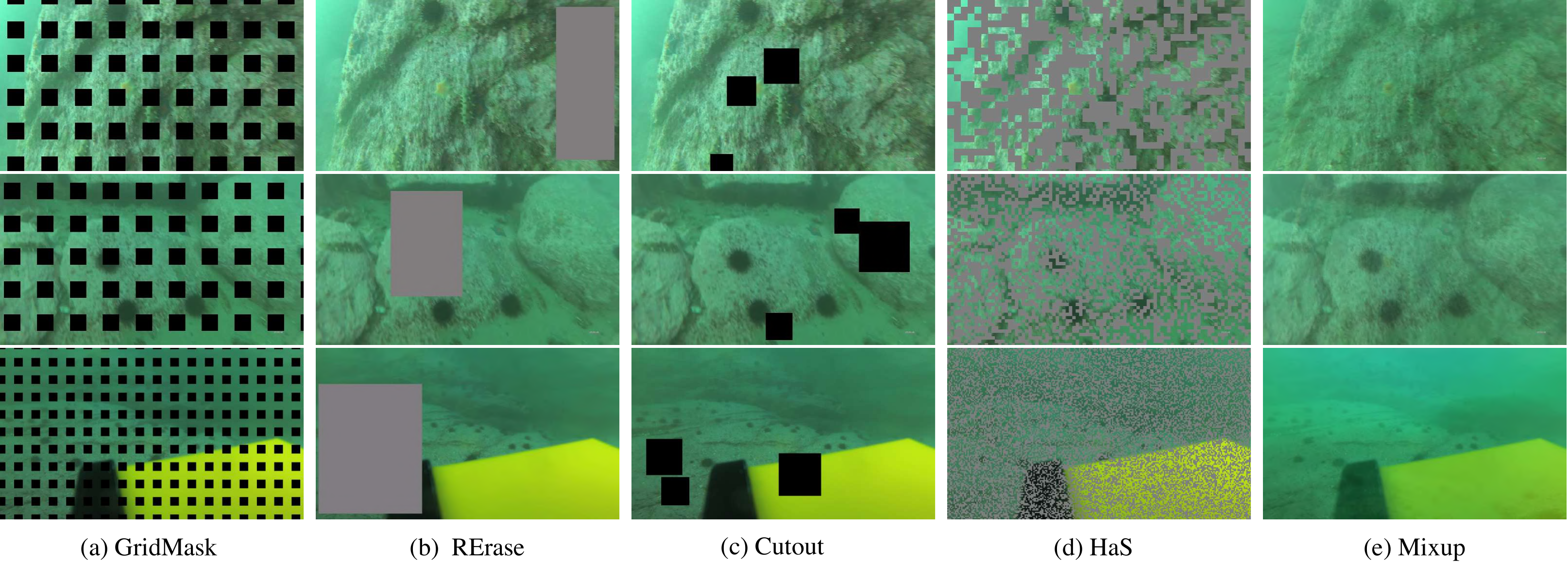}
\end{center}
   \caption{Examples of images processed by different augmentation methods. (a) GridMask \cite{Chen2020GridMaskDA}, (b) RErase \cite{Zhong2017RandomED}, (c) Cutout \cite{Devries2017ImprovedRO}, (d) HaS \cite{Singh2017HideandSeekFA}, (e) Mixup \cite{Zhang2017mixupBE}. GridMask, RErase, Cutout, and HaS drop image information in different ways and Mixup mixes up two images to take augmentation.}
\label{aug_comparison}
\end{figure*}
\begin{table}[!t]
\caption{Image Quality study results on UDD and URPC2018. AquaNet with pre-trained model is employed to study the image quality. +Poisson blending phase and +Lp (Learning phase) denote the dataset produced by corresponding phase is employed to train AquaNet.}
\centering 
\begin{tabular}{|l|c|c|}
\hline
&UDD (mAP$_{50}$)&URPC2018 (mAP$_{50}$)\\ 
\hline  
 AquaNet& 55.3 &  60.3\\ 
\hline
+Poisson blending phase& 64.5& 61.7\\ 
\hline
+Lp (w/o content loss)& 66.3& 62.1\\ 
\hline  
+Lp (w/o region loss)& 65.2& 61.5\\ 
\hline
+Lp (w/o adversarial loss)& 66.8& 62.4\\ 
\hline
+Lp & 72.1& 63.1\\ 
\hline
\end{tabular}
\label{t8}
\end{table}
\subsubsection{Image Quality Study}
As \cite{chen2020reveal} points out, color enhanced underwater images may not improve the performance of detectors although the enhanced images look much better than the origin images in human perspective. Poisson GAN is proposed to improve detectors' performance, so the quality study should be detector-oriented. Here, we use UDD and URPC2018 to study the image quality and the results are shown in Table \ref{t8}. URPC2018 contains 2,901 images for training and the testing set is not available. We select 414 images from the training set as the testing set (we choose one picture every seven in the total sequence). The augmented URPC2018 dataset has 18,531 images and the numbers of 3 creatures are 30,287, 110,262, 15,995. Notably, we use the same model parameters and the same cropped object set in AUDD to expend URPC2018. From Table \ref{t8}, we can find that datasets processed by Poisson blending phase achieve a large improvement compared with the origin dataset and the dataset processed by learning phase further improve the detector's performance significantly both on UDD and URPC2018. 
In addition, the DR loss contains three loss functions (\emph{i.e.,} content loss, region loss, and adversarial loss), and removing any loss function of DR loss during training Poisson GAN will result in worse image quality. 
Through this study, Poisson GAN is validated to have the capacity to improve the detector's performance. 
\subsubsection{Pre-training Dataset}
We conduct experiments to find out the performance of different initializations (Scratch, ImageNet pre-trained model, and the pre-trained model from pre-training dataset) in underwater open-sea farm object detection. YOLOv3 is trained for 70 epochs on pre-training dataset to get the pre-trained model. YOLOv3 is trained on AUDD with three different models and tested on the UDD test set as shown in Table \ref{t6}. 

Compared with the original result on UDD (see TABLE \ref{t3}), AUDD may solve the insufficient training data and class-imbalance problems in UDD. With our pre-trained model, YOLOv3 can be improved significantly from 46.8\% to 80.5\% on mAP$_{50}$, which is 11.7\% higher than the model trained from scratch and 9.0\% higher than the one trained with ImageNet. It shows that the pre-training dataset can provide a much better initialization for the detector in underwater open-sea farm object detection. 

\begin{table}[!t]
\caption{Comparisons with other GANs. PSNR denotes the Peak Signal to Noise Ratio. A higher score means the processed image is closer to the original one. 5,000 pictures generated from the Poisson blending phase are employed to calculate the PSNR.}
\centering 
\begin{tabular}{|l|c|}
\hline
Method& PSNR (dB)\\ 
\hline  
Poisson GAN& \bf 30.22\\
\hline
CycleGAN \cite{2017arXiv170310593Z}& 18.87\\
\hline
StarGAN \cite{DBLP:journals/corr/abs-1711-09020}& 24.72\\
\hline
UGATIT \cite{DBLP:journals/corr/abs-1907-10830}& 20.31\\
\hline
MUNIT \cite{huang2018munit}& 16.84\\
\hline  
\end{tabular}
\label{t10}
\end{table}

\subsubsection{Comparison With Image-to-Image Translation GANs}


GAN-based augmentation methods usually use image2image GANs to generate unreal images shown in Fig. \ref{comparison_2} to improve classifiers in the image classification task.
Therefore, we utilize some image2image GANs \cite{2017arXiv170310593Z,DBLP:journals/corr/abs-1711-09020,DBLP:journals/corr/abs-1907-10830,huang2018munit} to compare with our Poisson GAN.
TABLE \ref{t10} shows the Peak Signal to Noise Ratio (PSNR) scores (calculated form 5,000 pictures generated from the Poisson blending phase) of different methods from which Poisson GAN achieves the heighest score (30.22 dB) which means it could remain a better structure or texture of underwater scenes than others.
In addition, other methods will disappear some objects as shown in Fig. \ref{comparison_2}, which is harmful for training detectors and the class-imbalance problem is still in there after the transformation.
Hence, they cannot be used to generate an expanded dataset like AUDD to improve the detector empirically.

In contrast, Poisson GAN can change the position of objects and keep both the objects and underwater environment textures. The images generated by it are real enough to be the supplement of UDD.

\begin{table}[!t]
\renewcommand\tabcolsep{2.5pt}
\caption{Comparisons with different data augmentation methods. We show results without test augmentation / flip test for each method.}
\centering 
\begin{tabular}{|l|c|c|c|c|}
\hline
Method&mAP$_{50}$&mAP$_{cu}$&mAP$_{ur}$&mAP$_{sc}$\\ 
\hline  
Baseline & 55.3/58.1 & 38.9/44.1 & 80.1/81.5 & 47.0/48.6 \\ 
\hline
GridMask \cite{Chen2020GridMaskDA}&53.6/55.8 &37.5/39.3 &79.2/81.1 &44.0/46.9 \\ \hline
RErase \cite{Zhong2017RandomED}& 54.9/56.4 &40.1/44.7 &82.1/82.9 &42.5/41.5 \\ 
\hline
Cutout \cite{Devries2017ImprovedRO}& 54.3/56.3 & 39.7/44.2 & 81.8/82.9 & 41.5/41.6 \\ 
\hline
HaS \cite{Singh2017HideandSeekFA}& 30.7/30.9 & 9.1/9.1 & 65.7/67.9 & 17.2/15.7 \\ 
\hline
Mixup \cite{Zhang2017mixupBE}&55.1/60.6 &44.2/51.5 &80.7/81.7 &40.5/48.6 \\
\hline
Poisson GAN & {\bf 72.1}/\bf 80.8&{\bf 56.1}/\bf 64.7&{\bf 91.2}/\bf 94.7&{\bf 68.9}/\bf 82.8\\
\hline 
\end{tabular}
\label{t7}
\end{table}

\subsubsection{Comparison With Other Data Augmentation Methods}
We use some available augmentation methods \cite{Chen2020GridMaskDA,Zhong2017RandomED,Devries2017ImprovedRO,Singh2017HideandSeekFA, Zhang2017mixupBE} which can be used in general object detection to compare with our Poisson GAN. Examples are shown in Fig. \ref{aug_comparison}. GridMask, RErase, Cutout, and HaS drop image information in different ways and Mixup mixes up two images to take augmentation. In Table \ref{t7}, AquaNet with our pre-trained model is trained on AUDD in Poisson GAN and on UDD in other methods. All the models are tested on the UDD test set. Baseline means using random flip, random scaling (between 0.6 to 1.3), and cropping to train AquaNet. In GridMask, RErase, Cutout, and HaS, the corresponding method is first used before using the baseline augmentation mentioned above. In Mixup, the baseline augmentation is first used. The baseline augmentation is also employed in Poisson GAN. Finally, the image is normalized to around zero, and all the models are trained to convergence.

Among all the methods, information-dropping methods (\emph{i.e.}, GridMask, RErase, Cutout, HaS) perform worse than than the baseline method because dropping information could make the class-imbalance problem more serious. 
In the flip test, the accuracy of Mixup is improved by 2.5\% compared with Baseline, but there is still a large gap compared with the performance of Poisson GAN. Poisson GAN improves the detection accuracy significantly (22.7\% on mAP$_{50}$) compared with Baseline, and solves the class-imbalance problem to a great extent. Moreover, Poisson GAN could also use Mixup to further improve the performance and a combination of several augmentation techniques may further improve the performance; we do adopt the combination augmentation strategy in our engineering works.

\section{Conclusions}
In this paper, we propose an underwater open-sea farm object detection dataset (UDD) to promote the development of underwater robot grabbing. Then
the Poisson GAN is proposed to solve the class-imbalance problem in UDD by generating AUDD and the pre-training dataset. 
AquaNet with MFF block and MBP module is also designed to effectively detect mass small objects in underwater images. 
Finally, we conducted comprehensive experiments to verify the effectiveness of the proposed schemes. 
\section{Acknowledgments}
This work is supported in part by the National Natural Science Foundation of China (NSFC) under Grant 61976038, 61932020, and 61772108. All of the actions covered in this article have been approved by the Zhangzidao Group Co., Ltd. (the open-sea farm owner).

\bibliographystyle{IEEEtran}
\bibliography{egbib}

\begin{IEEEbiography}[{\includegraphics[width=1in,height=1.25in,clip,keepaspectratio]{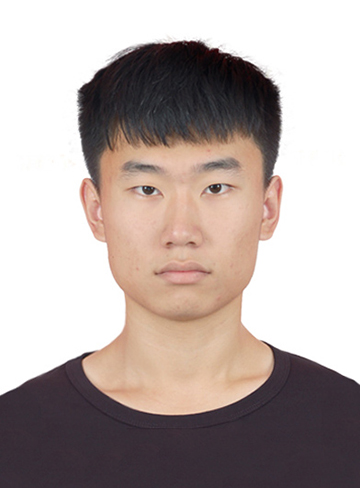}}]{Chongwei Liu}
received the B.E. and B.A. degrees in DUT-RU International School of Information Science \& Engineering in 2020 from the Dalian University of Technology, Dalian, China. He is currently a graduate student at the School of Software at the Dalian University of Technology in Liaoning, China. His research interests include image processing, computer vision, and deep learning.
\end{IEEEbiography}

\begin{IEEEbiography}[{\includegraphics[width=1in,height=1.25in,clip,keepaspectratio]{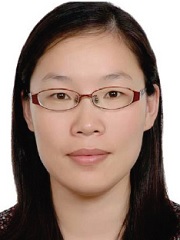}}]{Zhihui Wang}
is a Professor in the School of Software, Dalian University of Technology. Her research interests include social media computing and multimedia information retrieval. Dr. Wang received the B.E. and the Ph.D. degrees from the North Eastern University, Shenyang, and the Dalian University of Technology, Dalian in 2004 and 2010, respectively. Since November 2011, she has been a visiting scholar at the University of Washington.
\end{IEEEbiography}

\begin{IEEEbiography}[{\includegraphics[width=1in,height=1.25in,clip,keepaspectratio]{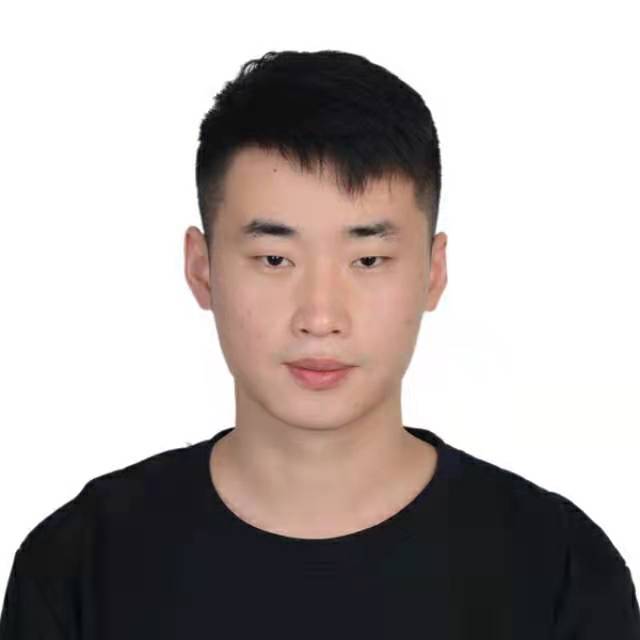}}]{Shijie Wang}
received the B.E. and M.S. degrees in the School of Software from the Dalian University of Technology, Dalian, China in 2017 and 2020, respectively. He is currently a doctoral candidate at the School of Software at the Dalian University of Technology in Liaoning, China. His research interests include image processing, computer vision, and deep learning.
\end{IEEEbiography}

\begin{IEEEbiography}[{\includegraphics[width=1in,height=1.25in,clip,keepaspectratio]{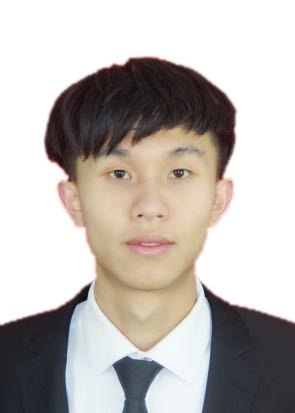}}]{Tao Tang}
received the B.E. degree in the School of Software in 2021 from the Dalian University of Technology, Dalian, China. He is currently a graduate student at the School of Intelligent Systems Engineering at Sun Yat-Sen University in Guangzhou, China. His research interests include image processing, computer vision, and deep learning.
\end{IEEEbiography}

\begin{IEEEbiography}[{\includegraphics[width=1in,height=1.25in,clip,keepaspectratio]{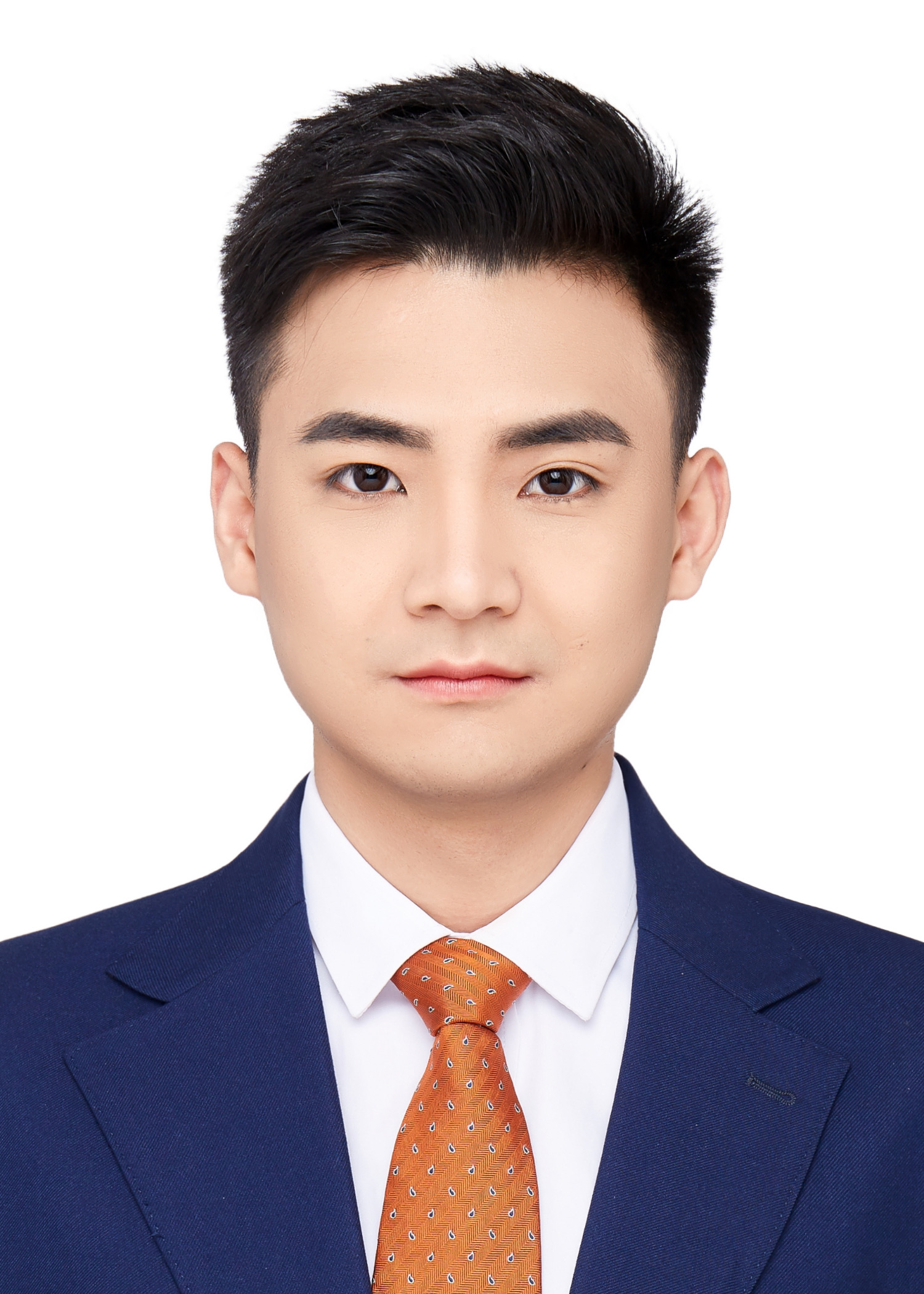}}]{Yulong Tao}
received the B.E. degree in the School of Ocean Science and Technology in 2019 from the Dalian University of Technology, Dalian, China. He is currently a graduate student at the State Key Laboratory of Information Engineering in Surveying, Mapping and Remote Sensing (LIESMARS) at Wuhan University in Wuhan, China. His research interests include image processing, computer vision, and deep learning.
\end{IEEEbiography}

\begin{IEEEbiography}[{\includegraphics[width=1in,height=1.25in,clip,keepaspectratio]{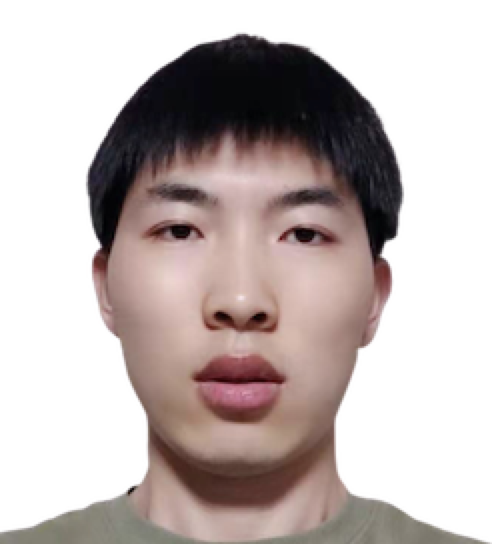}}]{Caifei Yang}
received the B.E. degree in the Northwest A\&F University, Yangling, China in 2018. He received the M.S. degree in the School of Software from the Dalian University of Technology, Dalian, China in 2020. His research interests include image processing, computer vision, and deep learning.
\end{IEEEbiography}

\begin{IEEEbiography}[{\includegraphics[width=1in,height=1.25in,clip,keepaspectratio]{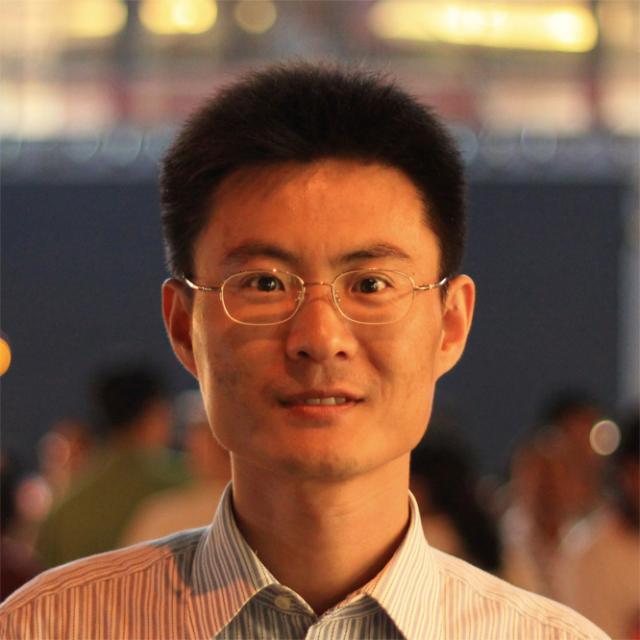}}]{Haojie Li}
is a Professor in the School of Software, Dalian University of Technology. His research interests include social media computing and multimedia information retrieval. Dr. Li received the B.E. and the Ph.D. degrees from Nankai University, Tianjin, and the Institute of Computing Technology, Chinese Academy of Sciences, Beijing, in 1996 and 2007 respectively. From 2007 to 2009, he was a Research Fellow in the School of Computing, National University of Singapore.\end{IEEEbiography}

\begin{IEEEbiography}[{\includegraphics[width=1in,height=1.25in,clip,keepaspectratio]{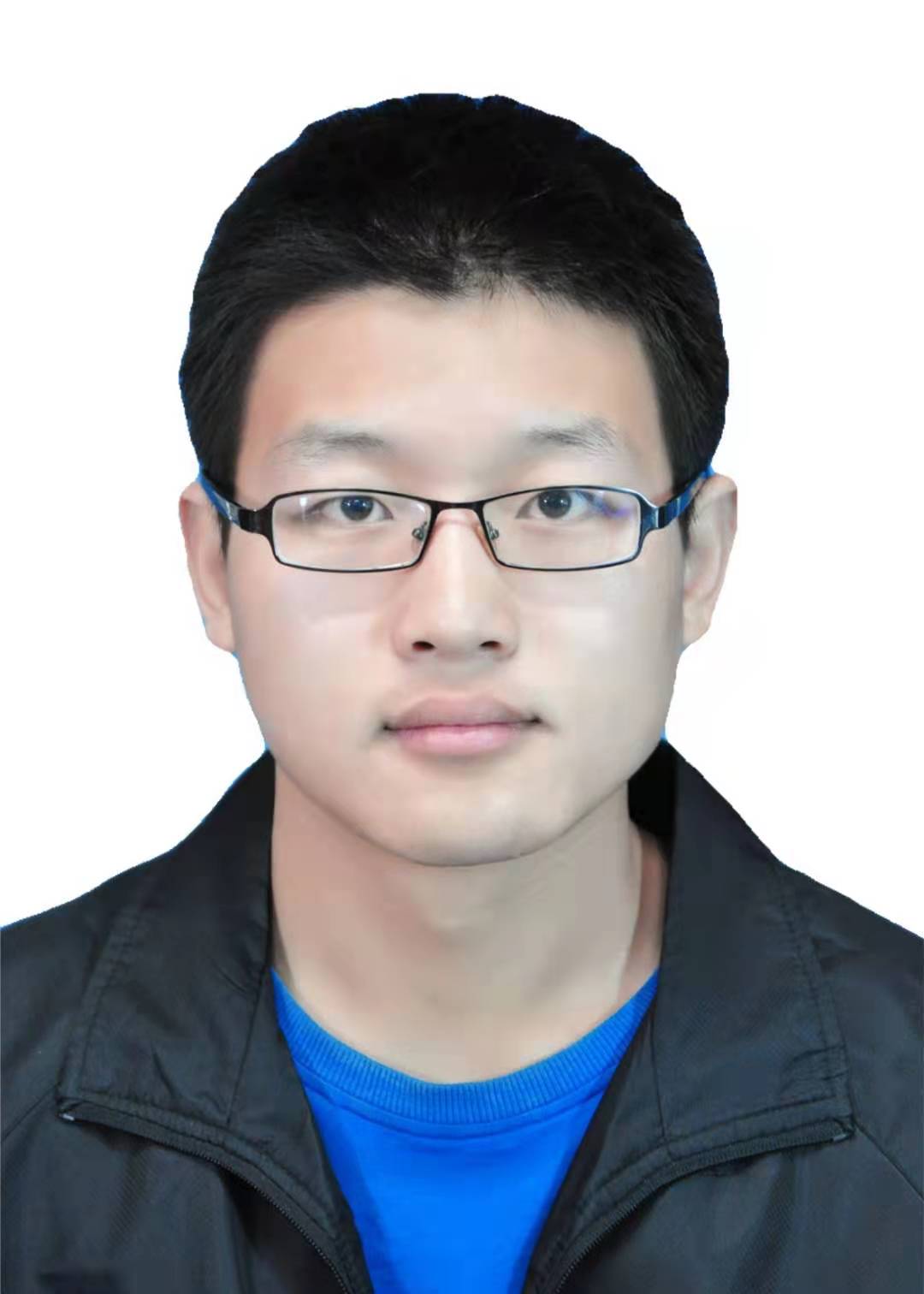}}]{Xing Liu}
received the M.S. degree in the School of Software from the Dalian University of Technology, Dalian, China in 2019. His research interests include image processing, computer vision, and deep learning.
\end{IEEEbiography}

\begin{IEEEbiography}[{\includegraphics[width=1in,height=1.25in,clip,keepaspectratio]{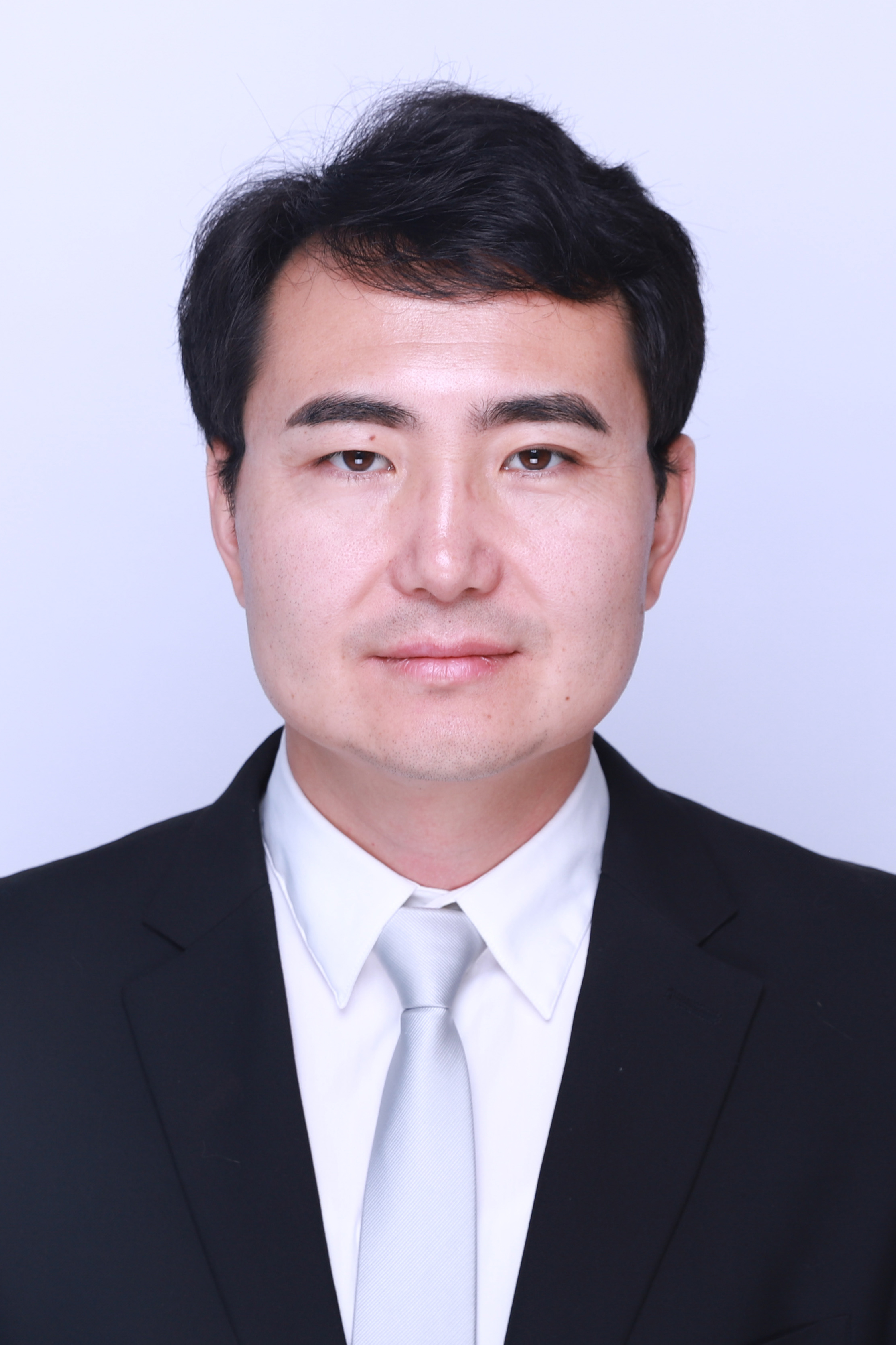}}]{Xin Fan (Senior Member, IEEE)} received the B.E. and Ph.D. degrees from Xi’an Jiaotong University, Xi’an, China, in 1998 and 2004, respectively. He was with Oklahoma State University, Stillwater, OK, USA, and the Southwestern Medical Center, The University of Texas at Dallas, Dallas, TX, USA, from 2006 to 2009, as a Post-Doctoral research fellow. He joined the Dalian University of Technology, Dalian, China, in 2009, where he is currently a full Professor. His current research interests include image processing and machine vision. Dr. Fan won the 2015 IEEE ICME Best Student Award as the Corresponding Author and two of his articles were selected as the Finalist of the Best Paper Award at ICME 2017.
\end{IEEEbiography}

\end{document}